\documentclass[10pt,twocolumn,letterpaper]{article}

\usepackage{iccv}
\usepackage{times}
\usepackage{epsfig}
\usepackage{graphicx}
\usepackage{amsmath}
\usepackage{amssymb}
\usepackage{cuted}
\usepackage{capt-of}
\usepackage{adjustbox}
\usepackage{comment}
\usepackage[accsupp]{axessibility}  

\newcommand{\ad}[1]{\textcolor{black}{#1}}

\newcommand{\lc}[1]{\textcolor{green}{#1}}

\newcommand{\md}[1]{\textcolor{yellow}{}}

\usepackage[pagebackref=true,breaklinks=true,letterpaper=true,colorlinks,bookmarks=false]{hyperref}

\iccvfinalcopy 

\def\iccvPaperID{9084} 

\ificcvfinal\fi

\begin{document}
\title{S2F2: Self-Supervised High Fidelity Face Reconstruction from Monocular Image}

\author
{\parbox{\textwidth}{Abdallah Dib\thanks{Equal contribution}\;\;
							   Junghyun Ahn$^{*}$\;\;                                
                                C\'edric Th\'ebault\;\;
                                Philippe-Henri Gosselin\;\;
                                Louis Chevallier
        }
        \\
        \\
{\parbox{\textwidth}{\centering InterDigital R\&I\;\;\;\;\;
       }
}
\vspace{-20px}
}
\maketitle
\ificcvfinal\thispagestyle{empty}\fi
\begin{strip}\centering
\includegraphics[width=\textwidth]{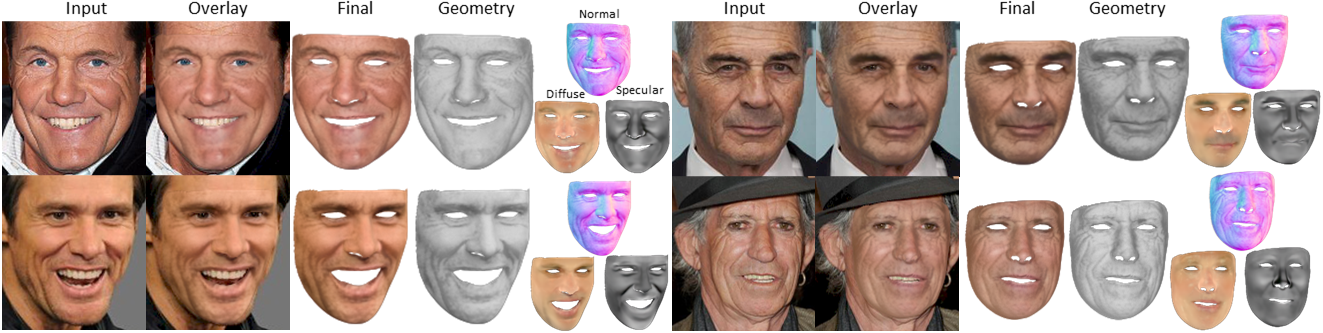}
\captionof{figure}{
Given a single image, our method achieves appealing 3D face reconstruction and estimates a dense detailed face geometry, spatially varying face reflectance (diffuse and specular albedos) and high frequency scene illumination.
\label{fig:teaser}}
\end{strip}
\begin{abstract}
We present a novel face reconstruction method capable of reconstructing detailed face geometry, spatially varying face reflectance from a single monocular image. We build our work upon the recent advances of DNN-based auto-encoders with differentiable ray tracing image formation, trained in self-supervised manner. While providing the advantage of learning-based approaches and real-time reconstruction, the latter methods lacked fidelity. In this work, we achieve, for the first time, high fidelity face reconstruction using self-supervised learning only. 
Our novel coarse-to-fine deep architecture allows us to solve the challenging problem of decoupling face reflectance from geometry using a single image, at high computational speed. Compared to state-of-the-art methods, our method achieves more visually appealing reconstruction.
\end{abstract}

\section{Introduction}
\label{sec:intro}
Fast, robust and high fidelity 3D face reconstruction has a wide range of applications in many domains such as interactive face editing,
video-conferencing, 
XR, Metaverse applications,
and visual effects for movies post-production.
Several approaches such as \cite{beeler2011high,wu2011shading,valgaerts2012lightweight,ghosh2011multiview,gotardo2018practical,riviere2020single} achieve high fidelity reconstruction, but require complex hardware setups (multi-view, lightstage). They are therefore not easily usable in most of the aforementioned applications.
Significant progress was made to achieve  high quality reconstruction from monocular image/video using optimization-based frameworks. Such methods \cite{Garrido2013,suwajanakorn2014total,Garrido:2016,dib2021practical} are generally slow, of limited robustness and not suitable for interactive scenarios. Also, their performances in challenging conditions (non-uniform lighting, extreme poses) are limited. 

Deep-based analysis-by-synthesis approaches have been investigated to leverage the generalization capabilities of machine learning. However, these methods \cite{tewari17MoFA,tewari18fml,tran2018nonlinear,tran2019towards} generally sacrifice reconstruction quality. Methods combining CNNs with differentiable rendering trained in a self-supervised manner have been introduced by Tewari~\textit{et al.} \cite{tewari17MoFA,tewari18fml,tewari2019fml}. These methods directly regress the parameters of a statistical morphable model and achieve real-time performance but fall short of the quality and fidelity because their estimated geometry and reflectance are bound by the statistical prior space which limits its generalization regarding the real diversity of face geometry and reflectance.

More recently, many works aim to improve the realism and fidelity of deep-based  methods by capturing either detailed geometry or reflectance but not both, which we discuss next. First, to capture detailed geometry, and because of the complexity of the problem, several methods rely on ground truth dataset obtained either from multi-view reconstruction setup (and/or lightstage) \cite{yamaguchi2018high,chen2019photo,lattas2020avatarme,yang2020facescape}, from synthetic data~\cite{sela2017unrestricted,sengupta2018sfsnet} or from a mixture of both~\cite{zeng2019df2net,abrevaya2020cross}.  Feng~\textit{et al.} \cite{feng2021learning} is the only self-supervised method that captures detailed geometry. However, this method only captures medium-scale geometry details and misses high-frequency geometry variations. Additionally, its estimated reflectance is restricted to the statistical prior space which limits its generalization regarding the real diversity in face geometry and reflectance.  Second, and to improve the reflectance,  Dib~\textit{et al.} \cite{dib2021towards} combined ray tracing and self-supervised learning to capture medium-scale reflectance details. However, this method restricts the estimated geometry to a parametric face model preventing high-frequency facial details (such as wrinkles, folds...) to be captured.  To our knowledge, there is no existing self-supervised methods that can jointly  estimate detailed geometry and reflectance. 

The first contribution of  this work, is the introduction of  the  first self-supervised method that  jointly estimate detailed geometry and reflectance. This is accomplished via our novel coarse-to-fine architecture, with an adapted training strategy which allows our method to efficiently solve the ambiguous and complex problem of separating detailed geometry from reflectance from a single image taken under uncontrolled lighting conditions.

The second contribution, is the combination, for the first time, of differentiable ray tracing with vertex-based renderer at training time  to overcome the problem of edge discontinuities of the ray tracing. This allow our method to benefit from both renderers. On one hand, ray tracing accurately models self-shadows and on the second hand, the vertex-based renderer evaluates correctly the whole geometry including boundaries. This leads to a significant improvement in the estimated geometry compared to Dib~\textit{et al.} \cite{dib2021towards} that uses only ray tracing.  

Finally, the aforementioned contributions enable to take a big leap forward in reconstruction quality for self-supervised methods and lead to superior face reconstruction when compared to recent state-of-the art methods. To our knowledge, this is the first time a self-supervised method reaches this level of fidelity and realism. Our robust face attributes estimation (diffuse, specular and normal) leads to practical applications such as face attribute editing and relighting.

\section{Related works}
\label{sec:sota}
Methods such as \cite{beeler2011high,valgaerts2012lightweight,ghosh2011multiview,gotardo2018practical} deliver high-fidelity face reconstruction from multi-view or light-stage setup, but they are generally expensive and not applicable for in-the-wild conditions (many cameras, specific lighting). 
In this work, we  are interested in face reconstruction/tracking methods that only use image or video as input and do not require any external hardware setup beyond the camera. These methods can be split into two categories: optimization-based and learning-based approaches.
\vspace{4px}
\\
\textbf{Geometry and reflectance modeling}
Statistical 3D Morphable Models - 3DMM - is the main building block for a wide range of optimization-based and learning-based methods ~\cite{blanz1999morphable,li2017learning,Egger20Years,zollhoefer2018facestar}. This statistical model adds a lot of structure and priors to face reconstruction problem from monocular image or video and makes it tractable. However, due to the low-dimensional space of 3DMM, subject specific medium and high frequency geometry and albedo details cannot be modeled.
Additionally, the skin reflectance model of 3DMM can only model the diffuse albedo and may bake shadows/specularity in the albedo. \cite{smith2020morphable} proposes a drop-in replacement for the basic lambertian reflectance model of 3DMM incorporating a diffuse and specular priors. In this work, we base our reconstruction on the 3DMM geometry with the statistical diffuse and specular prior of \cite{smith2020morphable} and we train a novel multi-stage deep network to capture fine diffuse and geometry details.

Most of optimization-based methods like \cite{Garrido:2016,Garrido2013,suwajanakorn2014total,wu2016anatomically,andrus2020face,dib2021practical,thies2016face2face,gerig2018morphable} rely on the same 3DMM parametrization, they provide generally precise reconstruction at the expense of a high computation cost and are sensitive to difficult lighting conditions.

Among the learning-based methods, deep convolution neural networks (CNN)
are
effective at
direct face reconstruction~\cite{tran2018nonlinear,tran2019towards,tewari17MoFA,tewari18fml,tewari2019fml,Koizumi20Look,sengupta2018sfsnet,deng2019accurate,ploumpis2020towards,sanyal2019learning,tu2019joint,3ddfa_cleardusk,guo2020towards}. Tewari~\textit{et al.} \cite{tewari17MoFA} proposed the first self-supervised autoencoder-like method to estimate face attributes based on 3DMM. 
The advantage of these self-supervised methods is that they can be trained on large corpus of unlabeled images. However they generally fall short of  reconstruction precision
because of their simplified underlying scene parameterization (pure-Lambertian BRDF to model skin reflectance and low-order spherical harmonics to model light). Their inability to model self-shadows is also a
possible
reason for their instability under challenging lighting conditions. 
Dib~\textit{et al.} \cite{dib2021towards} proposes a self-supervised method that significantly improves over these methods and solves many of these limitations. For instance, 
it uses a cook-torrance BRDF to model skin reflectance, and captures personalized albedos outside the statistical prior space. It also uses a differentiable ray tracing to model self-shadows. However the reconstructed geometry of their method is still limited by 3DMM space. 
\vspace{4px}
\\
\textbf{Detailed geometry reconstruction}
Capturing fine detailed geometry on top of global face structure is a pre-condition to achieve high fidelity face reconstruction. However, because of the complexity of the problem, methods such as \cite{riviere2020single} uses specific hardware setup which is not applicable in-the-wild.
Others, such Cao~\textit{et al.} \cite{cao2015real} relies on ground truth dense data \cite{beeler2011high}, or data captured using a lightstage (or multi-view) such as \cite{yang2020facescape,bagautdinov2018modeling,huynh2018mesoscopic,yamaguchi2018high,chen2019photo,lattas2020avatarme}. However, acquiring such ground truth data is not always possible. 

Optimization based methods like \cite{jiang20183d,Garrido2013,Garrido:2016} use shape-from-shading \cite{zhang1999shape} to capture fine geometry details. However these methods may not generalize well and are computationally expensive.

Some deep-based methods rely partially on synthetic data (\cite{sela2017unrestricted,richardson2017learning,zeng2019df2net}) or a mix of labeled and unlabeled data \cite{abrevaya2020cross} to capture fine detailed geometry.
The bias introduced by 
these methods may impede the resulting precision due to the mismatch with real data distribution. They also do not estimate face reflectance.  Sengupta~\textit{et al.} \cite{sengupta2018sfsnet} uses synthetic data to train their network which  estimates a normal map and skin reflectance (limited to pure-lambertian BRDF) but does not capture high frequency geometry details.
More recently, Feng~\textit{et al.} \cite{feng2021learning}  learns an 'expression-dependent' displacement model in-the-wild and is the only method that relies only on unlabeled images for end-to-end training. However this method only captures medium-frequency displacement map and their estimated reflectance is restricted to the statistical albedo prior space which limits their reconstruction quality.
To our knowledge, our method is the first self-supervised method that jointly estimates: geometry at high frequency, spatially varying personalized skin reflectance with diffuse and specular albedos and high frequency illumination from a single monocular image. 
\vspace{4px}
\\
\textbf{Differentiable rendering}
Differentiable rendering is a key block in the context of analysis-by-rendering and several implementations exist. Tewari~\textit{et al.} \cite{tewari17MoFA} proposed an efficient vertex-based differentiable rendering that can only handle pure Lambertian BRDF and cannot capture self-shadows.
Works such as \cite{lyu2021efficient} propose a differentiable shadow computation method for this type of renderer.

Dib~\textit{et al.} \cite{dib2019face,dib2021practical} introduced a method which uses differentiable ray tracing for face reconstruction within a classic optimization framework. 
The key advantage of ray tracing over vertex-based renderer is the capacity of ray tracing to handle self-shadows where a visibility mask is calculated for each surface point with respect to each light during direct illumination pass. However, differentiable ray tracing is computationally expensive and memory consuming. 
Recently, Dib~\textit{et al.} \cite{dib2021towards} uses differentiable ray tracing in conjunction with a deep neural architecture for direct face reconstruction. In this scheme, inference does not incur the speed penalty of ray tracing and delivers near real-time performance with robust reconstruction in challenging lighting conditions.
Regarding the optimization process, a limitation of ray tracing is the noise on gradients originating at the objects boundaries as they are sampled by very few points. 
Solutions exist but remain expensive (\cite{Loubet2019Reparameterizing,Li:2018:DMC}). In this work, we combine a vertex-based renderer with a ray tracing renderer together with a deep neural architecture that takes advantage of both. On  one hand, ray tracing accurately models self-shadows, and on the second hand, the vertex-based renderer evaluates the whole geometry including the edges.
\section{Method}
\label{sec:method}
\begin{figure*}[t]
\includegraphics[width=\linewidth]{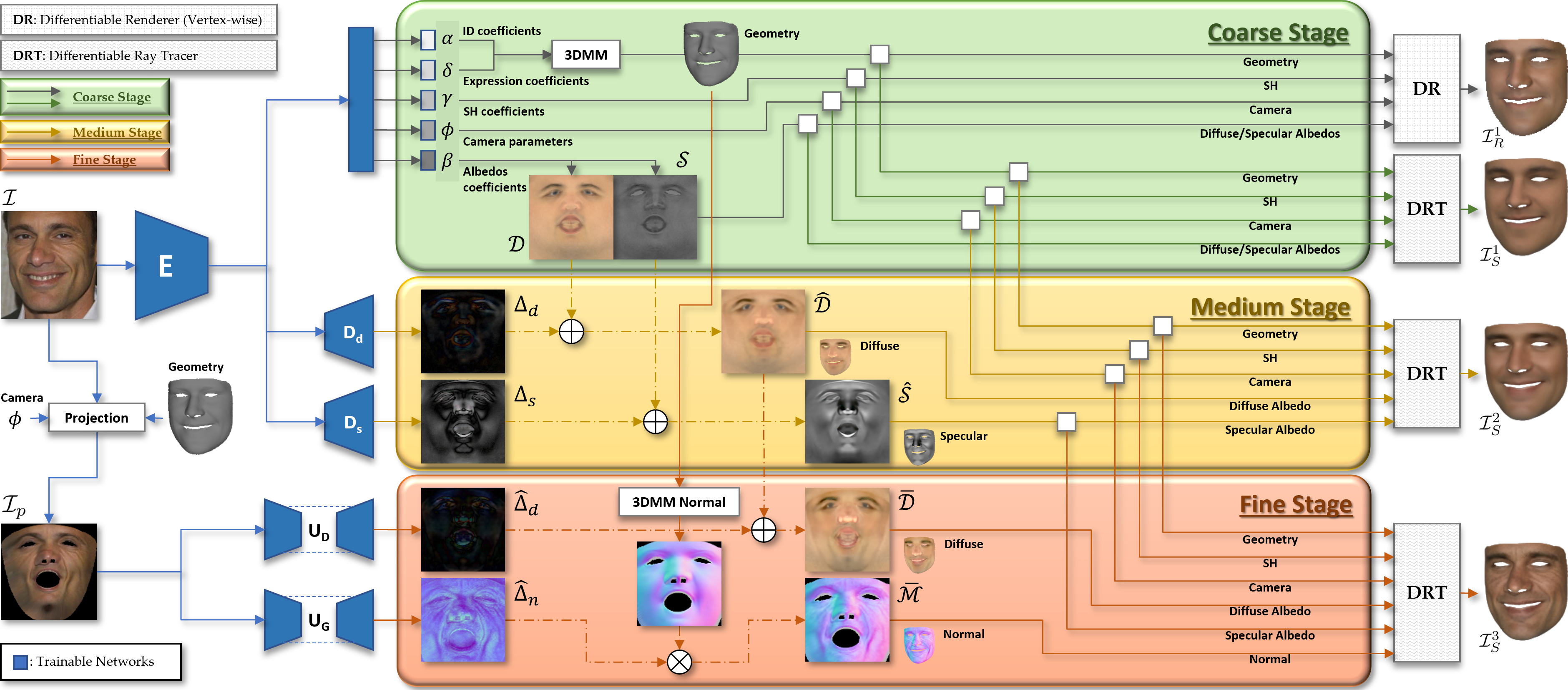}
  \caption{Our network architecture, trained end-to-end in a self-supervised manner, estimates face attributes (reflectance and detailed geometry) in a coarse-to-fine fashion (refer to section \ref{sec:method}).}
  \label{fig:overview}
\end{figure*}
Our goal is to obtain a high fidelity face reconstruction with faithful separation between reflectance and geometry attributes. To solve this ill-posed problem, we propose a novel multi-stage deep architecture, wherein different stages progressively refine the reconstruction.

Our network architecture, depicted in Figure \ref{fig:overview}, is composed of 3 stages denoted: 'Coarse', 'Medium' and 'Fine'. The 'Coarse' reconstruction relies on the statistical geometry and albedo priors space.
This base reconstruction lacks important geometry and albedo (diffuse and specular) details because it is restricted by the low dimensional space of the underlying model. The 'Medium' stage improves the previously estimated albedos without the limitations of the statistical prior.
Finally, the 'Fine' stage 
adds diffuse albedo and 
fine geometry details.

In the next, we introduce the scene attributes used by our formulation and we describe the network architecture. 
\subsection{Scene attributes}
\textbf{Face geometry}
Shape identity is modeled using \cite{blanz1999morphable,gerig2018morphable}'s statistical face model, and is given by $\mathsf{e} = \mathsf{a}_s + \mathsf\Sigma_s \mathsf{\alpha}$. $\mathsf{e}$ is a vector of face geometry vertices  with $N$ vertices. The identity-shape space is spanned by $\mathsf\Sigma_s \in \mathbb{R}^{3N \times K_s}$ composed of $K_s = 80$ principal components of this space. $\mathsf{\alpha} \in \mathbb{R}^{K_s}$ weights each coefficient of the 3DMM and $\mathsf{a}_s\in\mathbb{R}^{3N}$ is the mean face mesh vertices. We use linear blendshapes to represent face expressions over the neutral identity $\mathsf{e}$: $\mathsf{v} = \mathsf{e} + \mathsf\Sigma_e \mathsf{\delta}$, where $\mathsf{v}$ is the final vertex position displaced from $\mathsf{e}$ by blendshape weights vector $\mathsf{\delta} \in \mathbb{R}^{ K_e}$ and $\mathsf\Sigma_e \in \mathbb{R}^{3N \times K_e}$ composed of $K_e = 75$ components of the expression space.
\\
\textbf{Face reflectance}
Similar to \cite{dib2021towards}, we use a simplified Cook-Torrance BRDF~\cite{cook1982reflectance,walter2007Microfacet} with a constant roughness term. This BRDF model has the advantage of modelling 
specular reflections compared to the pure Lambertian BRDF. For each vertex, we define a diffuse $\mathsf{c}_i \in \mathbb{R}^3$ and a specular $\mathsf{s}_i\in \mathbb{R}^3$ albedos. The statistical diffuse  $\mathsf{c}\in\mathbb{R}^{3N}$ and specular $\mathsf{s}\in\mathbb{R}^{3N}$ albedos are obtained from  statistical prior of \cite{smith2020morphable}, where $\mathsf{c} = \mathsf{a}_r + \mathsf\Sigma_r \mathsf\beta$ and $\mathsf{s} = \mathsf{a}_b + \mathsf\Sigma_b \mathsf\beta$ with $\mathsf\Sigma_r, \mathsf\Sigma_b \in\mathbb{R}^{3N \times K_r}$ are the PCA for diffuse and specular reflectance with $K_r=80$. 
$\mathsf{a}_r$ and $\mathsf{a}_b$ are the average skin diffuse and specular reflectance.
We use the same coefficient $\beta$ to sample the diffuse and specular albedos as in \cite{smith2020morphable}.
\\
\textbf{Illumination}
Similar to \cite{dib2021towards}, we use nine spherical harmonics (SH) bands to model light. Dib~\textit{et al.} \cite{dib2021practical} showed that this high order SH parameterization provides a better light and shadows estimation when used with ray tracing compared to the widely used low-order 3 SH bands. An environment map of $64 \times 64$ is derived from SH to use with ray tracing. We define $\gamma \in \mathbb{R}^{9 \times 9 \times 3}$ the light coefficients.
\\
\textbf{Camera} We use the pinhole camera model and define $\phi = \{ \mathsf{T}, \mathsf{R}\}$ as the camera parameters with rotation $\mathsf{R} \in \mathsf{SO}(3)$  and translation $\mathsf{T} \in \mathbb{R}^3$.
\subsection{Coarse stage}
On Figure \ref{fig:overview}, the network \textbf{E} projects the input image $\mathcal{I}$ into the latent scene representation followed by a fully connected layer that predicts the semantic attributes vector $\chi = \{\mathsf\alpha, \mathsf\delta, \mathsf\phi, \mathsf\gamma, \mathsf\beta\}$.
Diffuse $\mathcal{D}$ and specular $\mathcal{S}$ texture representations are derived from $\beta$. These parameters are fed to a differentiable ray tracer (DRT) and to a vertex-wise differentiable renderer (DR) to generate two images $\mathcal{I}_S^1$ and $\mathcal{I}_R^1$ respectively. The following loss function is minimized during training:
\begin{equation}
 \label{eq:loss1}
 \mathsf{E}_{d}(\chi) + \mathsf{E}_{p}(\mathsf\alpha, \mathsf\beta) + \mathsf{E}_{b}(\mathsf\delta),
\end{equation}
where $\mathsf{E}_{d}$ is the data term equal to: 
\begin{equation}
 \label{eq:lossData}
\mathsf{E}_{d}(\chi) = \mathsf{E}_{ph}^S(\chi) +  w_{dr} \mathsf{E}_{ph}^R(\chi) +  w_{lm} \mathsf{E}_{land}(\chi),
\end{equation}
with $\mathsf{E}_{ph}^S$ is the pixel-wise photo-consistency loss between
input and ray traced pixels $\mathsf{p}_i, \mathsf{p}_i^S\in\mathbb{R}^3$:

\begin{equation}
 \label{eq:energyRayTrace}
\mathsf{E}_{ph}^S(\chi) =\sum_{i} |\mathsf{p}_i^S(\chi) - \mathsf{p}_i|,
\end{equation}
where  $\mathsf{p}_i^S=\mathcal{F}(\chi)$, with $\mathcal{F}$, the Monte Carlo estimator of the rendering equation \cite{Kajiya86Rendering}.
$\mathsf{E}_{ph}^R$ is the vertex-based photo-consistency loss between the projected mesh and the corresponding pixels in the input image, defined as follows:
\begin{equation}
 \label{eq:energyRaster}
\mathsf{E}_{ph}^R(\chi) =\sum_{i = 1}^{N} |\mathcal{B}(\mathsf{n_i}, \mathsf{c_i}, \mathsf{R_i}) - \mathcal{I} ( \Pi \circ \mathsf{C(\mathsf{v}_i}))|,
\end{equation}
where $N$ is the number of vertices,  $\mathsf{C(\mathsf{v_i}})$ is the projection of vertex $\mathsf{v_i}$ in the real image, equal to: $\mathsf{R}^{-1} (\mathsf{v_i} - \mathsf{T})$. $\Pi$ is the perspective camera matrix that projects a 3D vertex to a 2D pixel.
 $\mathcal{B}$ is the final irradiance equal to the sum of the diffuse and specular terms weighted by the specular intensity $s_i$ (details on $\mathcal{B}$ in supplementary material section A).
$\mathsf{E}_{land}$ is the landmark loss, which measures the distance between the $L = 68$ predicted facial landmarks and the projection of their corresponding vertex on the input image. These landmarks are obtained using the landmarks detector of \cite{bulat2017far}.  $\mathsf{E}_{p}$ is the statistical prior that regularizes against implausible face geometry and reflectance~\cite{dib2021practical}. $\mathsf{E}_{b}(\mathsf\delta)$ is a soft-box constraint that maintains $\mathsf\delta$ in the range $[0, 1]$.
\\
\textbf{Edge discontinuities}
Ray tracing can naturally models self-shadows by building a visibility mask for each surface point with respect to each light.
However, the major drawback of differentiable ray tracing is the discontinuities along geometry edges (\cite{Loubet2019Reparameterizing,Li:2018:DMC}). In fact, when solving for the rendering equation via Monte Carlo ray tracing \cite{Kajiya86Rendering}, very few points are sampled on these areas. As a result, back-propagation fails to handle geometry edges accurately during the optimization.
Several solutions have been proposed to overcome this limitation but they are generally very computationally expensive (\cite{lyu2021efficient,Li:2018:DMC}). For instance, \cite{Li:2018:DMC} explicitly samples the geometry edges, which extremely penalizes the training time as it needs to calculate the silhouette edges of the geometry.
While landmarks are mainly used to guide the training, in the particular case of ray tracing they can help mitigating the aforementioned limitation. However the geometry is not as precise as it could be.
As an efficient solution, we introduce in this work a new loss term $\mathsf{E}_{ph}^R$ (eq. \ref{eq:energyRaster}) which relies on a vertex-based differentiable renderer. This leads to a more accurate reconstruction, by taking advantage of ray tracing (which can model self-shadows) and vertex-based rendering (for better gradients on geometry edges) without significant cost. For instance, it only takes $370$ ms to process (forward-backward) an image using our method compared to $42$ seconds for the method of \cite{Li:2018:DMC}.
\subsection{Medium stage}
\label{ssec:medium}
The albedos (diffuse and specular) and geometry estimated by the previous stage are bound by the statistical prior space and can only capture low frequency components of the skin reflectance and geometry. Our goal is to obtain personalized
albedos (outside this space) with detailed geometry. Estimating these parameters jointly in a self-supervised manner is challenging. For this, we proceed with a coarse-to-fine strategy and we start by capturing personalized medium diffuse and specular albedos outside the statistical prior space. The challenge here is to avoid mixing diffuse and specular albedos and also to avoid baking unexplained shadows in these albedos. For this, we use the same technique of Dib~\textit{et al.} \cite{dib2021towards} which estimates personalized shadow-free albedos. For this, we train two additional decoders, $\mathbf{D_d}$ and $\mathbf{D_s}$, in a self-supervised way, that estimate diffuse $\Delta_d$ and specular $\Delta_s$ increments to be added on top of the previously estimated textures, $\mathcal{D}$ and $\mathcal{S}$, respectively. The resultant textures, $\mathcal{\hat{D}}$ and $\mathcal{\hat{S}}$, are used to generate a new image $\mathcal{I}_S^2$. 
We note that the second stage has only access to latent space of \textbf{E} allows this stage to focus on separating medium diffuse from specular albedo without worrying about high frequency geometry details that are discarded naturally by design.
We define $\hat{\chi} = \{\mathsf\alpha, \mathsf\delta, \mathsf\phi, \mathsf\gamma, \mathcal{\hat{D}}, \mathcal{\hat{S}}\}$ and we minimize the following loss function:
\begin{align}
\label{eq:loss2}
\mathsf{E}_{d}(\hat{\chi}) +
\mathsf{E}_{sc}(\mathcal{\hat{A}},\mathcal{A}) + 
 w_{m} \mathsf{E}_{m}(\mathcal{\hat{A}}) +  w_b \mathsf{E}_{b}(\mathcal{\hat{A}}),
\end{align}
where $\mathsf{E}_{sc}(\mathcal{\hat{A}},\mathcal{A})=w_s \mathsf{E}_{s}(\mathcal{\hat{A}}) +w_{c} \mathsf{E}_{c}(\mathcal{\hat{A}},\mathcal{A})$, and $\mathcal{A}$ is either $\mathcal{D}$ or $\mathcal{S}$. $\mathsf{E}_{s}$ and $\mathsf{E}_{c}$ are the symmetry and consistency regularizers (similar to \cite{dib2021towards}) used to avoid baking residual shadows in the personalized albedos.
$\mathsf{E}_{m}$ is a constraint term that ensures local smoothness at each vertex, with respect to its first ring neighbors. 
\subsection{Fine stage}
While the previous stage allows to obtain more personalized diffuse and specular albedos, these albedos remain generally blurry 
and still miss details. Also the geometry is restricted to the low-dimensional space of  3DMM. For this, we leverage the U-net based architecture (with skip connections) which are very efficient at capturing these fine details.
First, using the pose and the geometry produced by the first stage, we project the input image $\mathcal{I}$ in the uv-space. This projection, denoted as $\mathcal{I}_p$, is passed to two U-net networks, $\mathbf{U_G}$ and $\mathbf{U_D}$. $\mathbf{U_G}$ predicts a normal map $\hat{\Delta}_n$ used to displace the original normal vectors of the coarse mesh. We denote $\mathcal{\Bar{M}}$ the final normal map used for shading, where each vector $\mathsf{\Bar{m}}_i$ in $\mathcal{\Bar{M}}$ equal to:  $\mathsf{\Bar{m}}_i = \mathsf{T}_i \bigotimes \mathsf{\Bar{n}}_i$,
with $\mathsf{\Bar{n}}_i$ sampled from $\hat{\Delta}_n$ and $\mathsf{T}_i$ is a column-wise matrix that stack the original normal $\mathsf{n}_i$, tangent  $\mathsf{t}_i$ and bi-tangent $\mathsf{b}_i$ vectors in the camera coordinate system (more on normal mapping with ray tracing in \cite{veach1998robust}). $\mathbf{U_D}$ predicts an increment $\hat{\Delta}_d$ that is added to the estimated diffuse albedo $\mathcal{\hat{D}}$ of the previous stage. We denote $\mathcal{\Bar{D}}$ as the resulting texture. 

We define $\Bar{\chi} = \{\mathsf\alpha, \mathsf\delta, \mathsf\phi, \mathsf\gamma, \mathcal{\Bar{M}}, \mathcal{\Bar{D}}\}$ and we train $\mathbf{U_G}$ and $\mathbf{U_D}$  in a self-supervised manner by minimizing the following loss function: 
\begin{align}
\label{eq:loss3}
 \mathsf{E}_{d}(\Bar{\chi}) + 
 \mathsf{E}_{sc}(\mathcal{\Bar{D}}, \mathcal{\hat{D}}) +
 w_{m}^f \mathsf{E}_{m}(\mathcal{\Bar{A}}) +  w_{b}^f \mathsf{E}_{b}(\mathcal{\Bar{A}}),
\end{align}
where $\mathsf{E}_{sc}(\mathcal{\Bar{D}},\mathcal{\hat{D}})=w_{s}^f \mathsf{E}_{s}(\mathcal{\Bar{D}}) +w_{c}^f \mathsf{E}_{c}(\mathcal{\Bar{D}},\mathcal{\hat{D}})$, and $\mathcal{\Bar{A}}$ is either $\mathcal{\Bar{D}}$ or $\mathcal{\Bar{M}}$.
The regularization terms $\mathsf{E}_{s}$ and $\mathsf{E}_{c}$ play an important role in avoiding baking unexplained shadows in diffuse texture $\mathcal{\Bar{D}}$ in case 
our lighting model did not recover the light correctly.
Also these regularizers contribute in producing a good separation between diffuse and geometry details. However, they sacrifice some albedo details (please refer to the ablation section \ref{sec:ablation}). Finally, we note that we experimented using an additional U-net to capture a specular increment $\hat{\Delta}_s$ (similar to the diffuse) but we did not obtain substantial improvements in the reconstruction quality.  
%
\subsection{Training strategy}
We proceed with the following training strategy. We first train $\mathbf{E}$  (with the fully-connected layer) for 30 epochs, in a non-supervised manner, to directly regress $\chi$ by minimizing eq.~\ref{eq:loss1}.
Next, we train 
$\mathbf{D_d}$, $\mathbf{D_s}$, 
and $\mathbf{E}$ for 10 epochs while minimizing the loss in eq.~\ref{eq:loss2}. We follow the same training strategy proposed by \cite{dib2021towards} to separate diffuse and specular albedos, which consists in starting with a high regularization value of $w_{c}$ for diffuse texture (in eq.~\ref{eq:loss2}), and then in progressively relaxing its value  during training to allow for the diffuse albedo to capture more details. 
Next, we fix 
$\mathbf{D_d}$ and $\mathbf{D_s}$, then 
we train $\mathbf{U_G}, \mathbf{U_D}$ and $\mathbf{E}$ for 5 epochs, to estimate a normal map and an enhanced diffuse map respectively,  by minimizing the photo-consistency loss in eq.~\ref{eq:loss3}.
To avoid over-fitting one component, and to obtain a plausible separation between these attributes, we start with a high weight for $w_{c}^f$ and we progressively relax this constraint to allow 
$\mathcal{\Bar{D}}$
to capture more albedo details.

Finally, we note that the vertex-based loss in eq.~\ref{eq:energyRaster} is used in the whole training process with the goal to assist and guide the pixel-wise photo-consistency loss of ray tracing (eq.~\ref{eq:energyRayTrace}) at different stages,  so to alleviate the problem of noisy edge gradients that ray tracing exhibits.
\begin{figure*}
\hspace{35pt}Input\hspace{35pt} Overlay\hspace{65pt} Final\hspace{35pt} Shading\hspace{15pt} Normal\hspace{15pt} Diffuse\hspace{15pt} Specular \\
\centering
\includegraphics[width=0.9\linewidth]{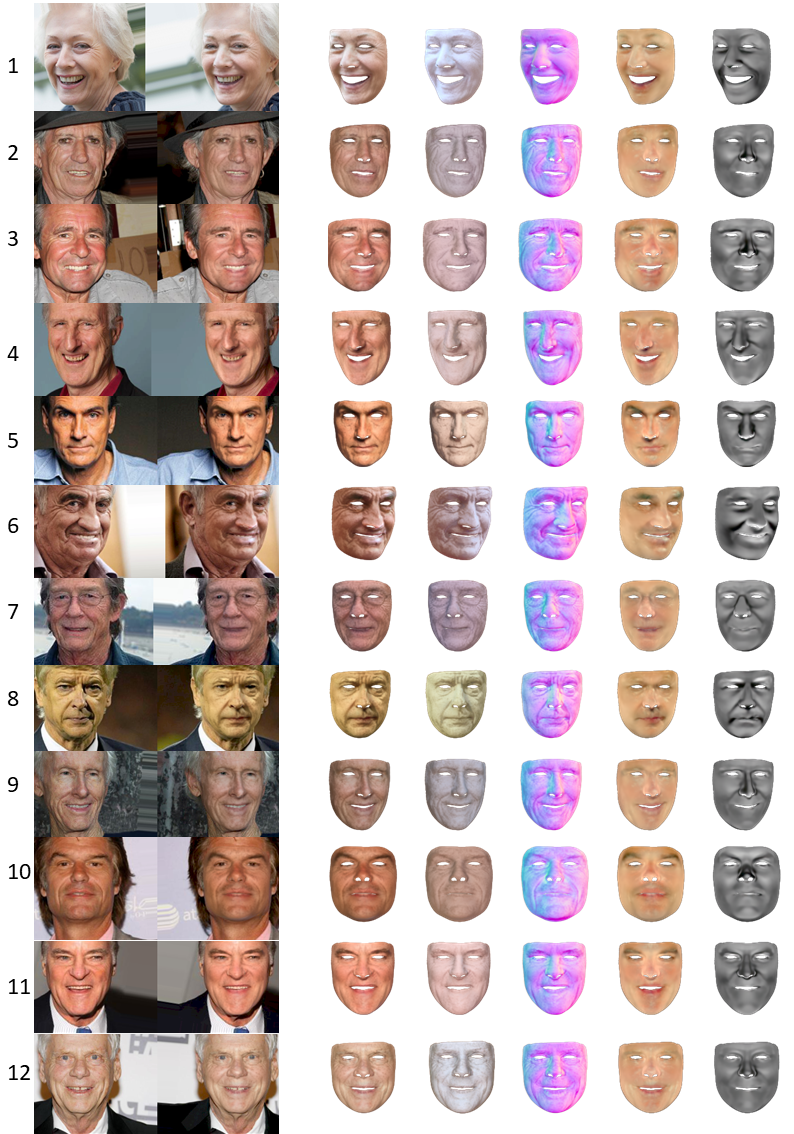}
  \caption{Results with challenging face details. From left to right: Input image, overlay of our final reconstruction on the input image, final reconstruction, shading, normal, diffuse and specular.}
  \label{fig:catOld1}
\end{figure*}
\section{Results}
\label{sec:results}
For training, we use CelebA dataset \cite{liu2015faceattributes} where images were cropped to $256\times256$.  Output textures of different networks have the same resolution as the input image. We implemented our network using PyTorch \cite{paszke2017automatic}. Ray tracing is based on the method of \cite{Li:2018:DMC}. During training, we use 8 samples per pixels for ray tracing. More implementation details are in supp. material, section B.

Figure \ref{fig:teaser}, Figure \ref{fig:catOld1} and supp. material (section E)  show successful face reconstruction of more than 100 subjects with challenging face details, extreme lighting conditions, challenging head pose/expression and different skin type (makeup, beards)... For all these subjects, our method successfully estimates personalized albedos and captures fine detailed geometry, which leads to appealing reconstruction at high fidelity. Even in challenging lighting conditions, our method successfully estimates shadow-free maps (diffuse, specular and normal). All this, at very high computational speed, where at inference, our method takes $131$ ms to process an image on a Nvidia RTX 2080 Ti. We note that while ray tracing penalizes our training time, it is not needed at inference time and our estimated attributes are compatible with existing rendering engines. Finally, our robust estimation of scene light, face reflectance and geometry provides explicit controls over the face attributes which leads to practical applications such as face attribute editing (aging) and relighting as shown in supp. material (section E). Finally, in supp. video, we also show reconstruction on video sequence.
%
\begin{figure}[t]
\centering
\includegraphics[width=\linewidth]{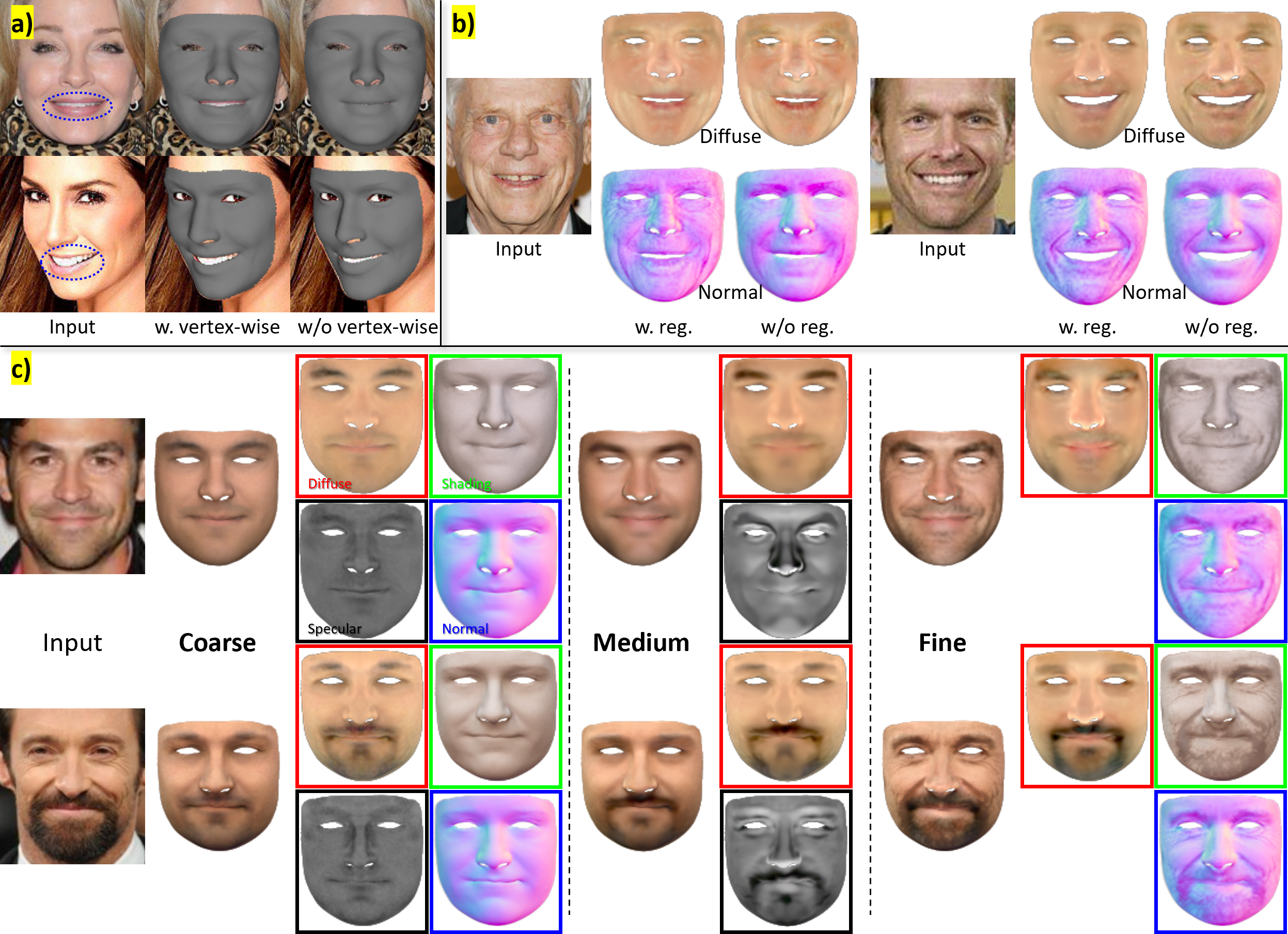}
  \caption{ablation studies (section \ref{sec:ablation})}
  \label{fig:ablation}
\end{figure}
\vspace{-5px}
\section{Ablation}
\label{sec:ablation}
\textbf{Importance of vertex-based renderer}
In this experiment, we study the importance of the vertex-based renderer to overcome the problem of noisy edge gradients of the ray tracer. 
For this, we trained \textbf{E} by dropping the vertex-based renderer based loss term (eq~\ref{eq:energyRaster}) from equation~\ref{eq:lossData}. We compare the estimated mesh to the one that use the full energy term (ray tracing + vertex-wise). 
The results in Figure \ref{fig:ablation} a) show that the estimated meshes using both the vertex-based renderer and ray tracer are more accurate than the ones obtained using ray tracing only (especially around the mouth edges). Quantitatively, we evaluate both methods on 100 subjects from Facescape dataset \cite{yang2020facescape} with various type of facial expressions. We compute the vertex position error with respect to ground truth mesh, and we obtain 2.288/1.671 mm (mean error/std deviation) for the 'vertex-based + ray tracing' compared to 2.831/1.782 mm for 'ray tracing only' which show that combining ray tracing with the vertex-based renderer improves the reconstructed geometry. We note that reason  why the improvement may look small is that 'vertex-based + ray tracing' aims to improve the geometry on very small area around the edges.
\\
\textbf{Regularization}
In this experiment, we study the importance of the symmetry and consistency regularizers ($\mathsf{E}_{sc}$) used in equation~\ref{eq:loss3} to separate the diffuse and geometric details faithfully. For this, we trained $\mathbf{U_G}$ and $\mathbf{U_D}$ by dropping these two regularizers. For both subjects in Figure \ref{fig:ablation} b), without these regularizers, some geometric details get baked in the albedo and leads to sub-optimal separation. Adding these regularizers produce more convincing separation.  While these regularizers play an important role in obtaining a correct separation between diffuse and geometry details, they sacrifice some albedo details.
%
\\
\textbf{Multi-Stage reconstruction}
In this experiment we show the importance of the 'Medium' and 'Fine' stages to improve the realism of the 'Coarse' reconstruction. Figure \ref{fig:ablation} c) shows the reconstruction obtained from the 'Coarse' stage  with the estimated statistical albedo priors. For the 'Medium' stage, we show the corresponding reconstruction and the enhanced diffuse and specular albedos. For the 'Fine' stage, we show the final reconstruction with the final diffuse and normal maps. We note that the diffuse albedo in 'Medium' stage is generally blurry and lacks some details and is significantly enhanced in the 'Fine' stage. Also, the detailed geometry captured in the 'Fine' stage significantly improves the realism of the final reconstruction. 
Quantitatively, we calculate the SSIM between the reconstruction of each stage and the original input image. On 1000 images, we obtain an average of $0.89/0.91/\textbf{0.97}$ for the coarse/medium/fine stages respectively (higher is better). This shows that the 'Fine' stage significantly improves the fidelity of the  reconstruction.
\begin{figure*}[t]
\centering
\includegraphics[width=1.\linewidth]{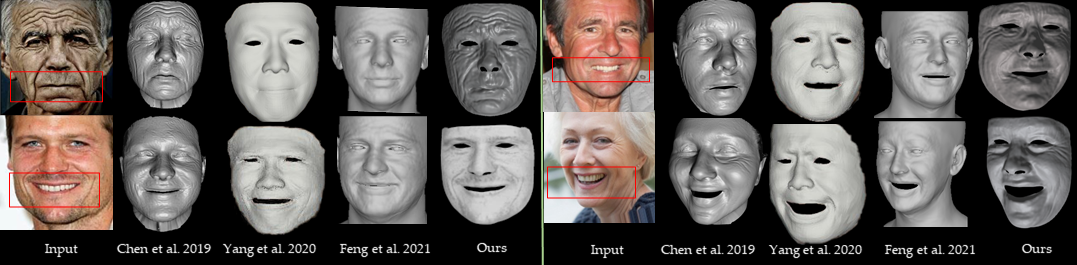}
  \caption{Comparison against Chen~\textit{et al.} \cite{chen2019photo}, Yang~\textit{et al.} \cite{yang2020facescape} and Feng~\textit{et al.} \cite{feng2021learning}
  }
  \label{fig:compChenFeng}
\end{figure*}
%
%

\vspace{-5px}
\section{Comparison}
\label{sec:comparison}
In this section, we compare, qualitatively and quantitatively,  our method to the state-of-the-art methods.
\subsection{Qualitative comparison}
Figure \ref{fig:compChenFeng} shows comparison against Chen~\textit{et al.} \cite{chen2019photo}, Yang~\textit{et al.} \cite{yang2020facescape} and Feng~\textit{et al.} \cite{feng2021learning}. For \cite{chen2019photo} and \cite{feng2021learning}, results are from authors' open implementation. For  \cite{yang2020facescape} results are generated  by the authors. The methods of \cite{chen2019photo} and \cite{yang2020facescape} use ground truth (GT) data to train their generative network while our method and \cite{feng2021learning} are  self-supervised methods and do not require any GT data. 
For all subjects, our method successfully capture most of geometry details especially for the top subjects that present challenging details. These details are barely captured or missed by other methods. Also our method shows better results on the wrinkles formed by the zygomaticus muscles (smile wrinkles around nose and mouth). Our method has a significantly better shape and expression recovery than all other methods, especially around the mouth (as highlighted in red rectangles), where the expression is incorrectly captured by other methods. 
\begin{figure}
\centering
\includegraphics[width=\linewidth]{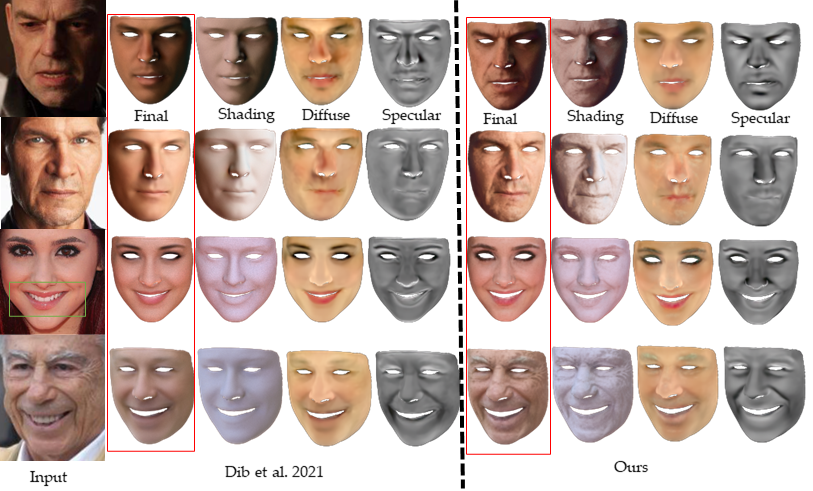}
  \caption{Comparisons against Dib~\textit{et al.} \cite{dib2021towards} (top subject from \cite{dib2021towards})
  }
  \label{fig:compDib}
\end{figure}
\\
Compared to Dib~\textit{et al.} \cite{dib2021towards} (Figure \ref{fig:compDib}), our method achieves robustness  against challenging lighting conditions similar to \cite{dib2021towards} and produces shadow free albedos (first two subjects). In addition, our method estimates better diffuse map and captures detailed geometry, while \cite{dib2021towards} restricts the geometry reconstruction to the low-dimensional space of 3DMM. This leads to a superior and high fidelity reconstruction of our method compared to \cite{dib2021towards} (highlighted in red rectangle). Finally, for third subject, our method that combine ray tracing and vertex-based rendering has a better expression recovery around the mouth than \cite{dib2021towards} that uses only ray tracing, which also confirms our earlier ablation study (highlighted in green rectangle).

\begin{figure*}
\centering
\includegraphics[width=.8\linewidth]{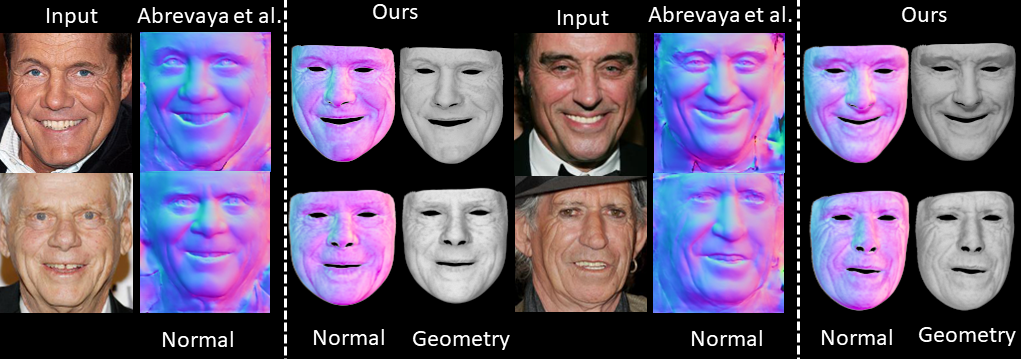}
  \caption{Comparison against Abrevaya~\textit{et al.} \cite{abrevaya2020cross}
  }
  \label{fig:compAbrevaya}
\end{figure*}

Figure~\ref{fig:compAbrevaya} show comparaison against the method of Abrevaya~\textit{et al.} \cite{abrevaya2020cross} on subjects with challenging facial details. The  method of Abrevaya~\textit{et al.} \cite{abrevaya2020cross} uses a combination of labeled and unlabeled data to train the network that predicts normal map of the face. It also produces a complete normal map for the entire head (including eyes and moth interior) while our method restricts the reconstruction to the frontal region of the face (without eyes and mouth interior). However, our method captures more facial details (especially arround the eyes) than \cite{abrevaya2020cross}. Finally we note that \cite{abrevaya2020cross} only predicts a normal map (in camera space and not in uv space), and other face attributes are not estimated. Our method, estimates rich face attributes maps in uv space (normal, diffuse and specular), face geometry and scene light. 
\\
More comparisons against other recent methods can be found in supp. material, section C.
%
%
%
%
%
%
%
\subsection{Quantitative comparison}
\begin{figure}[t]
\centering
\includegraphics[width=\linewidth]{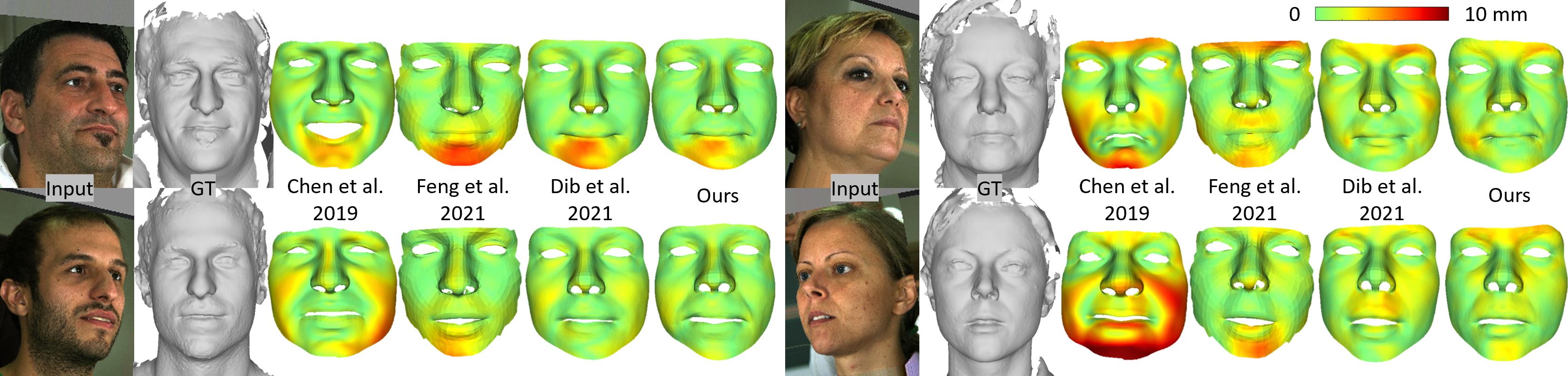}
  \caption{Vertex position error compared to GT mesh  of Superfaces dataset  \cite{berretti2012superfaces}. }
  \label{fig:meshEval}
\end{figure}

\textbf{Geometric evaluation} We first compare our estimated geometry to the state-of-the-art methods of Chen~\textit{et al.} \cite{chen2019photo}, Feng~\textit{et al.}~\cite{feng2021learning}, and Dib~\textit{et al.} \cite{dib2021towards} on the Superfaces dataset \cite{berretti2012superfaces} composed of 20 high resolution 3D ground truth (GT) face meshes (Table \ref{tab:posError} and Figure \ref{fig:meshEval}).
Table \ref{tab:posError} reports, for each method, on all subjects, the average error $\mu$ and standard deviation $\sigma$ for vertex position error. As shown in Figure~\ref{fig:meshEval}, the same mask is used for all methods. Feng~\textit{et al.}\cite{feng2021learning} achieves slightly better results than our method on average error while ours has a smaller standard deviation. Our method achieves better score than Chen~\textit{et al.} \cite{chen2019photo} and Dib~\textit{et al.} \cite{dib2021towards}. 
As depicted in Figure~\ref{fig:meshEval}, our approach measures lowest error around the mouth. Finally, we note that our method, which combines ray tracing with vertex-based renderer, has lower error than Dib~\textit{et al.}\cite{dib2021towards} that only uses ray tracing, which again confirms our earlier ablation study (Section \ref{sec:ablation}).
%
\begin{table}
\caption{Position error (mean/stdev) on Superfaces dataset \cite{berretti2012superfaces}}
\centering
\begin{adjustbox}{width=0.8\columnwidth}
\begin{tabular}{|l|c|c|c|c|}
\hline
 & Chen~\textit{et al.} \cite{chen2019photo} & Feng~\textit{et al.} \cite{feng2021learning} & Dib~\textit{et al.} \cite{dib2021towards}  & Ours  \\
\hline\hline
Mean err $\mu$    (mm) & 1.847 & \textbf{1.234} & 1.379  & 1.287 \\
Stdev $\sigma$ (mm) & 1.512 & 1.206 & 1.159 & \textbf{1.025} \\
\hline
\end{tabular}
\end{adjustbox}
\label{tab:posError}
\vspace{-5px}
\end{table}

\begin{figure}[ht]
\centering
\includegraphics[width=\linewidth]{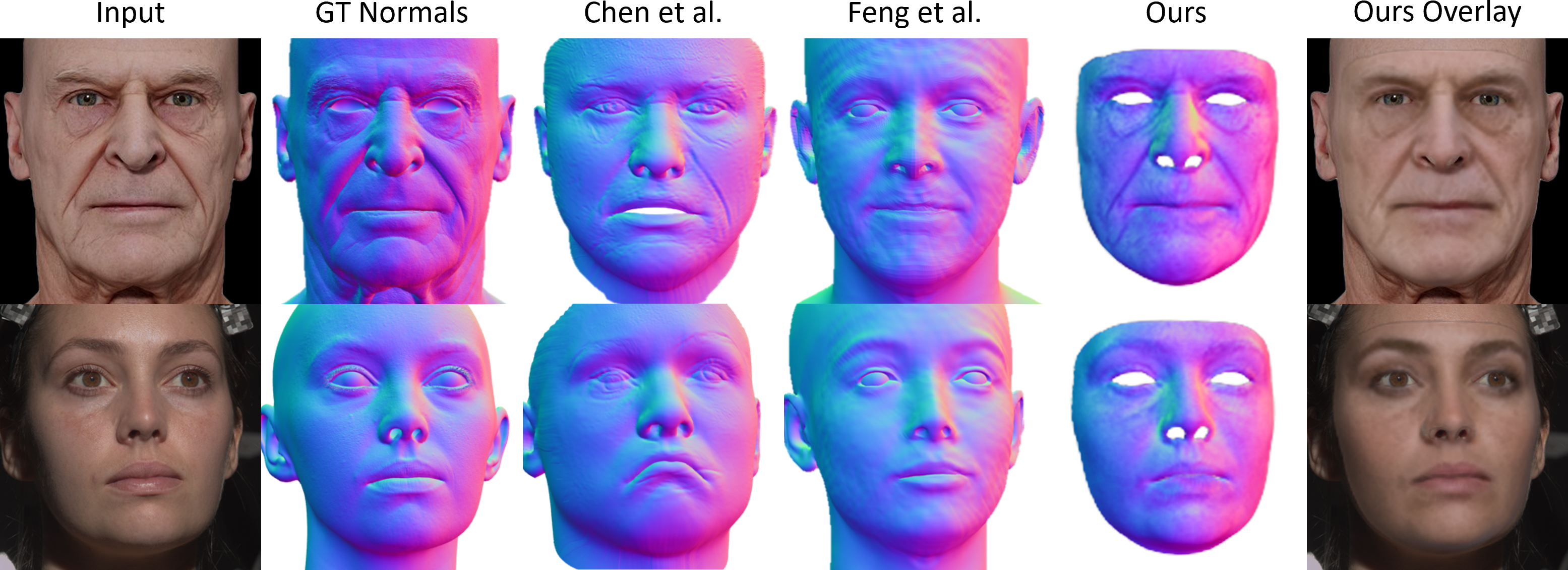}
  \caption{Estimated normal map compared to GT normal.
  }  
  \label{fig:normalEval}
\end{figure}

We also evaluate our method on the NoW dataset  \cite{RingNet:CVPR:2019}, which is composed of 80 subjects with a total of 1702 images. Results are reported in Table \ref{tab:nowPosError}. We note that this dataset only evaluates mesh in neutral pose, so expression accuracy is not evaluated. Nevertheless, our method achieves competitive results and a better score than Dib~\textit{et al.} \cite{dib2021towards}. Feng~\textit{et al.} \cite{feng2021learning} achieves the top score (more scores against other methods are on NoW website).
We note that for both datasets, for our method, the mesh used for comparison is the base 3DMM geometry estimated by the "coarse" stage as our method only estimates a normal map to model the finer details.
\vspace{-10px}
\begin{table}
\caption{Position error (median/mean/stdev) on Now dataset \cite{RingNet:CVPR:2019}}
\centering
\begin{adjustbox}{width=0.8\columnwidth}
\begin{tabular}{|l|c|c|c|}
\hline
 & Feng~\textit{et al.} \cite{feng2021learning} & Dib~\textit{et al.} \cite{dib2021towards}  & Ours  \\
\hline\hline
Median err (mm) & \textbf{1.09} & 1.26 & 1.24 \\
Mean err (mm) & \textbf{1.38} & 1.57  & 1.54 \\
Stdev (mm) & \textbf{1.18} & 1.31  & 1.29 \\
\hline
\end{tabular}
\end{adjustbox}
\label{tab:nowPosError}
\vspace{-5px}
\end{table}
%
%
%
%
%
%
\\
\vspace{4px}
\\
\textbf{Normal evaluation}
We also compare our estimated normal map to Chen~\textit{et al.} \cite{chen2019photo} and Feng~\textit{et al.} \cite{feng2021learning} on Emily \cite{emilyWikihuman} and on realistic 3D head model \cite{scanstore3D}  which we call 'Male014' in the next.
Results are reported in Figure \ref{fig:normalEval} and Table \ref{tab:normalError}. Since each method uses a different UV map parametrization, the comparison was done on the rendered image and not on the unwrapped texture using the same mask (as shown in Figure~\ref{fig:normalEval}).
For 'Male014', our method has the lowest mean error. For Emily, \cite{feng2021learning} has a better average error but our method has a lower standard deviation. Our method scores the best in error percentage under $20^{\circ}, 25^{\circ}$ and $30^{\circ}$ for both subjects except for Emily $20^{\circ}$. We note that the method of Chen~\textit{et al.} \cite{chen2019photo} does not correctly recover the mouth expression of both subjects (Figure \ref{fig:normalEval}). Finally, the last column in Figure \ref{fig:normalEval} shows the final reconstruction of our method overlaid on the input image.
%

\begin{table}
\caption{Mean angular error (degrees) and percentage of errors below $20^{\circ}, 25^{\circ}$ and $30^{\circ}$. (Top-row: \textbf{Male014}; Bottom-row: \textbf{Emily})}
\label{tab:normalError}
\centering
\begin{adjustbox}{width=0.8\columnwidth}
\begin{tabular}{|c|c|c|c|c|} 
\hline
Name & Mean$\pm$Std & $<$ 20$^{\circ}$ & $<$ 25$^{\circ}$ & $<$ 30$^{\circ}$ \\ 
\hline\hline
Ours & \textbf{22.9}$\pm$15.3 & \textbf{51.6}$\%$ & \textbf{67.0}$\%$ & \textbf{77.1}$\%$ \\ 
Feng et al. \cite{feng2021learning} & 24.2$\pm$14.5 & 43.8$\%$ & 59.7$\%$ & 72.4$\%$ \\ 
Chen et al. \cite{chen2019photo} & 26.2$\pm$15.6 & 39.4$\%$ & 55.8$\%$ & 68.5$\%$ \\ 
\hline
Ours & 16.8$\pm$ 9.9  & 71.2$\%$ & \textbf{84.7}$\%$ & \textbf{92.8}$\%$ \\
Feng et al. \cite{feng2021learning} & \textbf{15.2}$\pm$12.2  & \textbf{77.2}$\%$ & 84.5$\%$ & 89.4$\%$ \\
Chen et al. \cite{chen2019photo} & 17.0$\pm$13.9  & 72.4$\%$ & 82.7$\%$ & 88.2$\%$ \\
\hline
\end{tabular}
\end{adjustbox}
\end{table}
\vspace{-2px}
\section{Limitations, Future works and Conclusion}
\label{sec:conclusion}
\textbf{Limitations and Future works}
First, separating light color from skin color using a single image is not solved in this work and remains an open challenge, therefore, our method may sometimes bake some albedo color in the estimated light, that is why our shading may look reddish sometimes. 
Second, since we do not use symmetry and consistency regularizers for the normal map,
some shadows may get baked in the normal map when the method fails to recover to correct light of the scene. 
Our experiments show that adding these regularizers can mitigate this but sacrifices important geometry details. 
Third, regularizers used in eq.~\ref{eq:loss3} allow our method to obtain a correct separation between diffuse albedo and geometry details but this is at the expense of some albedo details (cf. section \ref{sec:ablation}). Finally, our method cannot handle occlusions, so face props (such as glasses) get baked in the estimated maps. 
Work such \cite{tran2018extreme} is important to tackle this.
As future work, our method does not connect expression to the geometry details as in \cite{feng2021learning} which is important to obtain realistic animated rigs. Such a function could leverage the rich representation produced by our pipeline.
Finally, our method estimates view and light dependent face attributes and can be extended to video/multi-view based reconstruction which can significantly improve the estimated facial attributes and alleviate a lot of ambiguity introduced by using a single image only. Some limitations are shown in supp. material, section D.
\\
\vspace{1px}
\\
\textbf{Conclusion}
In this work,  we push to a new level the principle of analysis-by-rendering within self-supervised learning framework by refining the modelling, the training as well as the rendering stages. First, by combining, for the first time, ray tracing with the vertex-based renderer at training time, to solve the problem of edge-discontinuities of ray tracing which significantly improves the overall geometry and suggests an improvement on the original ray tracing algorithm. Second, by introducing a coarse-to-fine deep architecture, with adapted training strategy, that solve, for the first time, the highly challenging problem of separating detailed face attributes (reflectance and geometry) from a single image, within a self-supervised setup, and under uncontrolled lighting conditions. Compared to recent state-of-the art, our method achieves superior reconstruction quality and produces more visually appealing results and define a new baseline for self-supervised monocular face reconstruction methods '\textit{in-the-wild}'.
\section{Acknowledgements}
We would like to thank authors in \cite{feng2021learning}, \cite{chen2019photo} and \cite{abrevaya2020cross} for sharing their code source publicly and authors of \cite{yang2020facescape} for processing our images with their method. 

{\small
\bibliographystyle{ieee_fullname}
\bibliography{ms}
}
\clearpage
\appendix
\section{Vertex-based implementation}
\label{sec:vwimpDetails}
Here we provide the implementation details for the final irradiance $\mathcal{B}$ used by the vertex-based renderer (please refer to eq. 4 in main document).

The final irradiance $\mathcal{B}$ is equal to the sum of the diffuse and specular terms weighted by the specular intensity $s_i \in\mathbb{R}$:
\begin{align}
    \mathcal{B}(\mathsf{n_i}, \mathsf{c_i}, \mathsf{R_i}) = (1 - \mathsf{s_i}) \cdot  \mathcal{B}_d(\mathsf{n_i}, \mathsf{c_i})  + \mathsf{s_i} \cdot \mathcal{B}_s(\mathsf{R_i}) 
\end{align}

The diffuse irradiance $\mathcal{B}_d$, is obtained by multiplying the SH coefficients, $B_{lm}$ of the light with SH coefficients ($A_{l}$) of the \textit{half-cosine} function (\cite{mahajan2007theory}):
\begin{align}
    \mathcal{B}_d(\mathsf{n_i}, \mathsf{c_i}) = \mathsf{c_i} \cdot \sum_{l = 0}^{8} \sum_{m = -l}^{l} A_{l} \cdot B_{lm} \cdot Y_{lm}(\mathsf{n_i})
\end{align}
with $\mathsf{c}_i\in\mathbb{R}^3$ and $\mathsf{n}_i\in\mathbb{R}^3$ are the diffuse albedo and normal vector for  each vertex respectively.

Similarly, The specular irradiance is  obtained using a spatial convolution of the SH light coefficients with the BRDF kernel of the roughness, which is constant in our simplified Cook-Torrance BRDF model:
\begin{align}
    \mathcal{B}_s(\mathsf{R_i}) = \sum_{l = 0}^{8} \sum_{m = -l}^{l} S_{l} \cdot B_{lm} \cdot Y_{lm}(\mathsf{R_i}),
\end{align}
with $\mathsf{R_i}$ is the reflection direction of the viewing vector $\mathsf{W_i}$ with respect to the surface normal, and $S_{l}$ are the SH coefficients of the BRDF function \cite{mahajan2007theory}.

\section{Implementation details}
\label{sec:impDetails}
We use  PyTorch \cite{paszke2017automatic} with Cuda-enabled backend gpu. Two GPUs were used for training with a total of 24GB of memory. We used Adam optimizer with default parameters. Celeba dataset \cite{liu2015faceattributes} is used for training with 3K images kept for validation. Images are aligned and cropped to a 256x256 resolution. Because ray tracing is generally slow, it takes 10 hours to do a single epoch. However, our method does not need ray tracing at test time and achieves fast inference ($131$ ms to process an image). We use 8 samples per pixel for ray tracing and a batch size of 8. For $\mathbf{E}$, we use a pre-trained \textit{ResNet-152}. The architecture for $\mathbf{D_d}$ and $\mathbf{D_s}$ is given in Table~\ref{tab:archD1D2}. For $\mathbf{U_G}$ and $\mathbf{U_D}$, we use a U-net with skip connections with a pre-trained \textit{vgg16} backbone from here\footnote{https://github.com/mkisantal/backboned-unet}. The last layer of    $\mathbf{U_D}$, $\mathbf{U_G}$, $\mathbf{D_d}$ and $\mathbf{D_s}$ is initialized to output zero increment at the beginning of the training. Landmarks weight $w_{lm} = 0.1$, vertex-based renderer weight $w_{dr} = 0.5$.  For the 'Medium' stage, smoothness regularizer $w_m = 0.0001$ for diffuse and specular albedos. The symmetry regularizer $w_s = 20$ for medium diffuse and specular albedos.  Consistency regularizer for specular albedo $w_c = 0.01$. For diffuse albedo, we start with a value of $0.2$ and we relax it by a factor of $2$ at each epoch. For the fine layer,  the smoothness regularizer $w_{m}^f = 0.0001$ for diffuse and normal maps. The symmetry regularizer $w{s}^f =10$ for the final diffuse. The consistency regularizer, we start with $w_{c}^f = 1.0$ and we relax it  by a factor of two at each epoch. 

\begin{table}
\centering
\resizebox{\columnwidth}{!}{
\begin{tabular}{|l|l|}
\hline
Layer & Architecture                                                            \\ 
\hline
1     & ct(256, 160, 3),bn=ELU,c2d(160, 256, 1),bn,ELU                          \\
2     & ct(256, 256, 3),bn,ELU=c2d(256, 128, 3),bn,ELU,c2d(128, 192, 3),bn,ELU  \\
3     & ct(192, 192, 3),bn,ELU,c2d(192, 96, 3),bn,ELU,c2d(96, 128, 3),bn,ELU    \\
4     & ct(128, 128, 3),bn,ELU,c2d(128, 64, 3),bn,ELU,c2d(64, 64, 3),bn,ELU     \\
5     & ct(64, 64, 3),bn,ELU,c2d(64, 32, 3),bn,ELU,c2d(32, 42, 3),bn,ELU        \\
6     & ct(42, 42, 3),bn,ELU,c2d(42, 21, 3),bn,ELU,c2d(21, 3, 3),Tanh           \\
\hline
\end{tabular}}
\caption{\label{tab:archD1D2}Architecture for $\mathbf{D_d}$ and $\mathbf{D_s}$ (ct: ConvTranspose2d, c2d: Conv2D, bn:BatchNorm2d (pytorch).}
\end{table}

\section{More qualitative comparison}
\label{sec:qualComp}
In this section, we show more qualitative comparison against state-of-the art methods. 

\begin{figure}
\centering
\includegraphics[width=\linewidth]{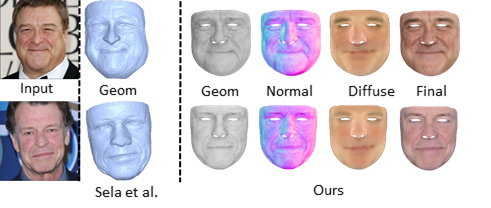}
  \caption{Comparison against Sela~\textit{et al.} \cite{sela2017unrestricted}. Subjects from authors' manuscript.}
  \label{fig:compSela}
\end{figure}

\begin{figure}[th]
\centering
\includegraphics[width=\linewidth]{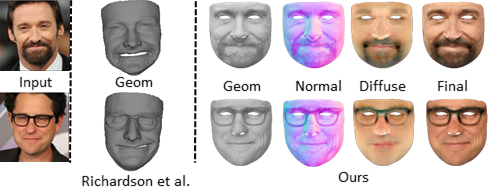}
  \caption{Comparison against Richardson~\textit{et al.} \cite{richardson2017learning}. Subjects from authors' manuscript.}
  \label{fig:compRichadrson}
\end{figure}

Figure~\ref{fig:compSela} and Figure~\ref{fig:compRichadrson} show comparison results against the method of Sela~\textit{et al.} \cite{sela2017unrestricted} and Richardson~\textit{et al.} \cite{richardson2017learning}. Subjects are taken from authors' manuscript. We note that these two methods only estimate detailed geometry while our method estimates geometry, reflectance and scene light.

\section{Limitations example}
\label{sec:limit}
\begin{figure}
\centering
\includegraphics[width=.9\linewidth]{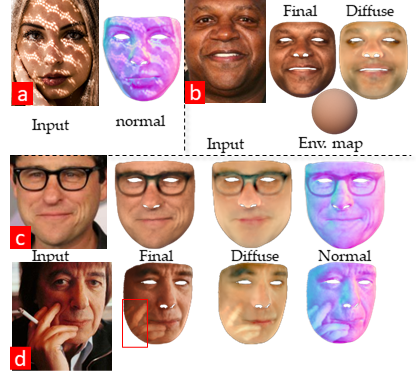}
  \caption{Limitations of our method.}
  \label{fig:limitation}
\end{figure}

Figure~\ref{fig:limitation} shows some limitations of our method. The description of the examples are as follows:
\begin{itemize}
  \item Subject a): Under external (foreign) shadows, our method fails to recover the light and leads to shadows baking in the estimated normal map. Tackling foreign shadows is a challenging problem, and the method such as Zhang~\textit{et al.} \cite{zhang2020portrait} described about the difficulties of handling this issue. We also note that our spherical harmonics based lighting model has its own limitations because it can only model infinite lights.
  \item Subject b): Separating light color from skin color using a single image remains challenging and our method does not solve for this. As a result, some skin color can get baked in the estimated light (represented here as an environment map). 
  \item Subject c) and d): Occlusions and face props can get baked in the estimated maps. In case of severe occlusions (subject d), our method may fail to correctly estimate the geometry (highlighted in red box).
\end{itemize}
%
%
%
%
\section{Results and relighting/aging applications }
\label{sec:results}
\begin{figure*}
\hspace{35pt}Input\hspace{35pt} Overlay\hspace{65pt} Final\hspace{35pt} Shading\hspace{15pt} Normal\hspace{15pt} Diffuse\hspace{15pt} Specular \\
\centering
\includegraphics[width=0.9\linewidth]{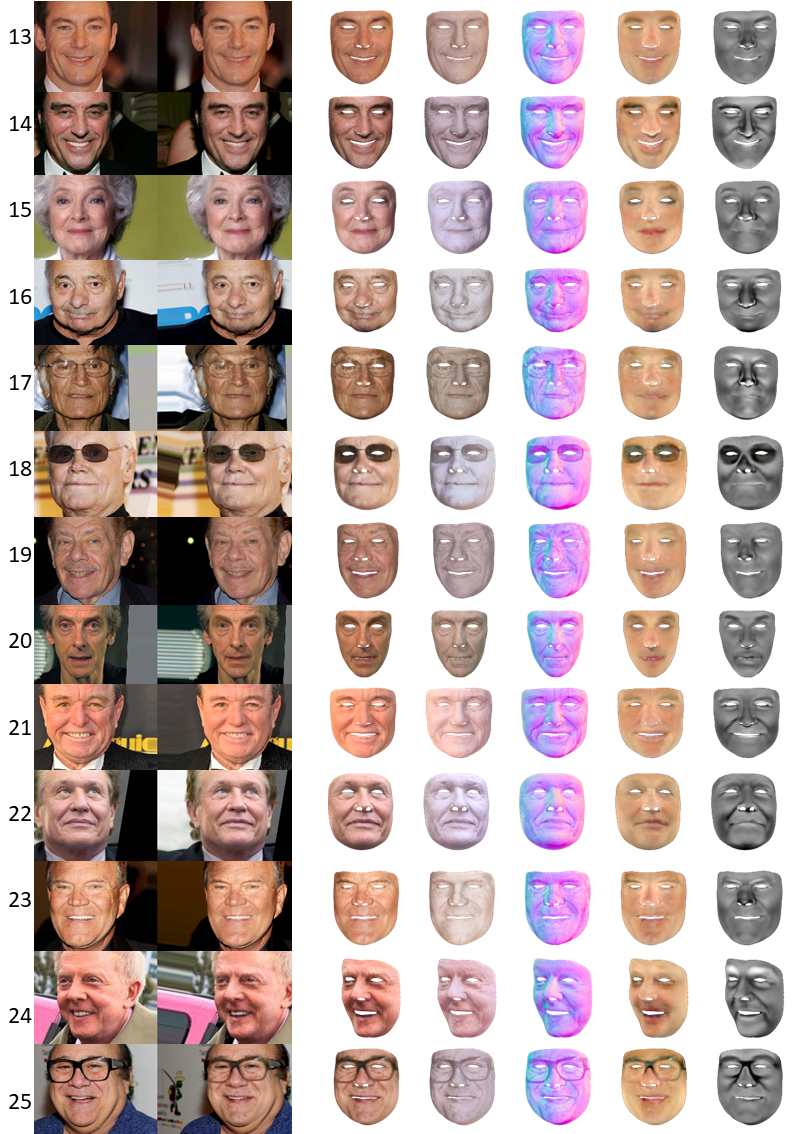}
  \caption{Results with challenging face details. From left to right: Input image, overlay of our final reconstruction on the input image, final reconstruction, shading, normal, diffuse and specular.}
  \label{fig:catOld2}
\end{figure*}

\begin{figure*}
\hspace{35pt}Input\hspace{35pt} Overlay\hspace{65pt} Final\hspace{35pt} Shading\hspace{15pt} Normal\hspace{15pt} Diffuse\hspace{15pt} Specular \\
\centering
\includegraphics[width=0.9\linewidth]{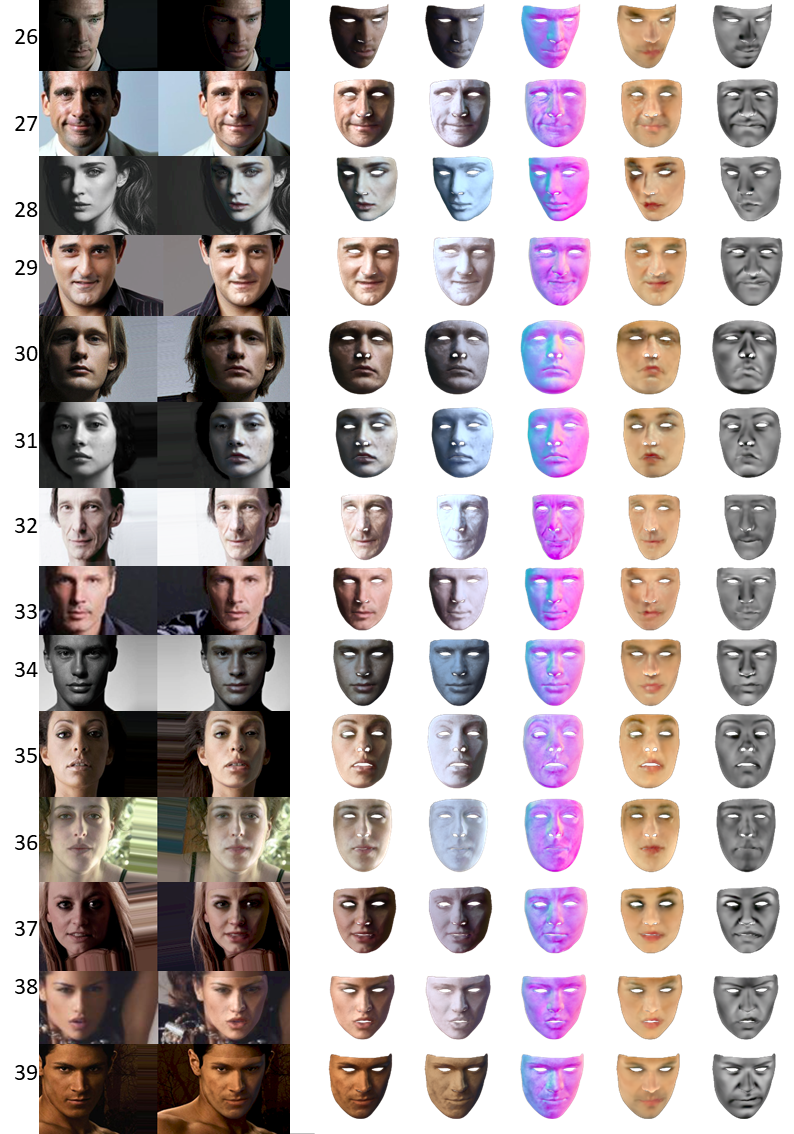}
  \caption{Results with challenging lighting conditions. From left to right: Input image, overlay of our final reconstruction on the input image, final reconstruction, shading, normal, diffuse and specular.}
  \label{fig:catLight}
\end{figure*}

\begin{figure*}
\hspace{35pt}Input\hspace{35pt} Overlay\hspace{65pt} Final\hspace{35pt} Shading\hspace{15pt} Normal\hspace{15pt} Diffuse\hspace{15pt} Specular \\
\centering
\includegraphics[width=0.9\linewidth]{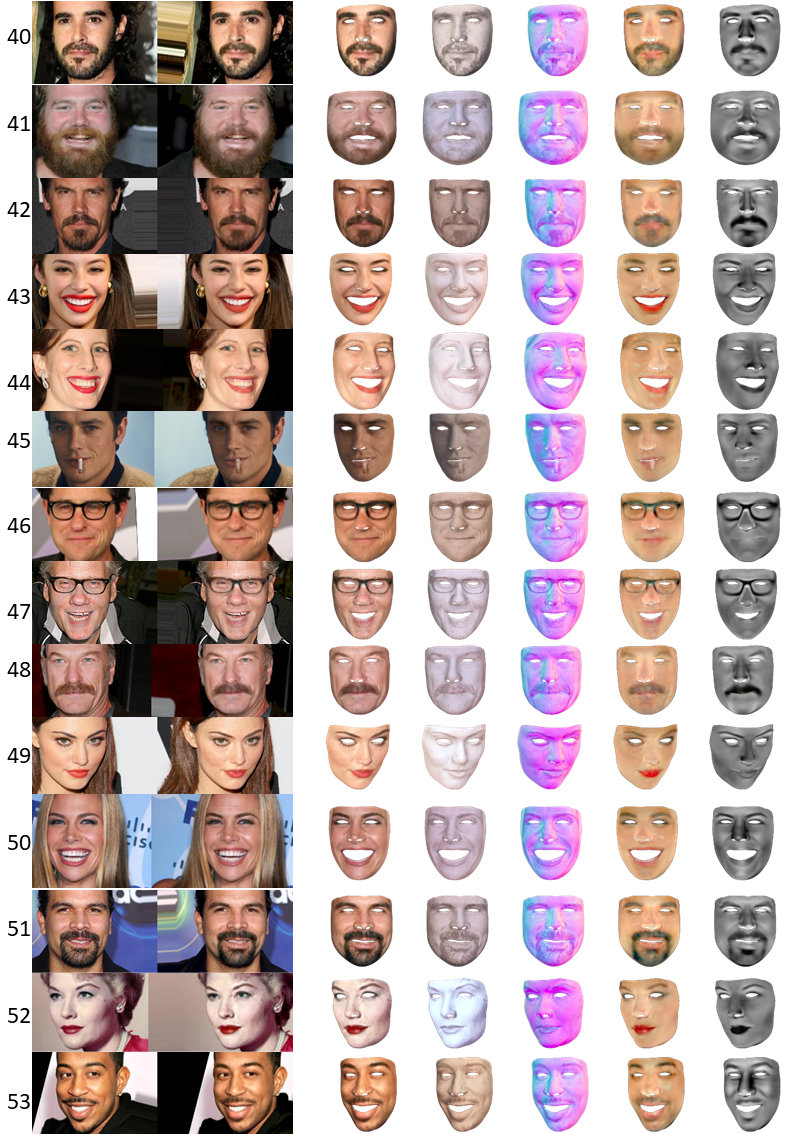}
  \caption{Results for subjects with make-up, beards and face props. From left to right: Input image, overlay of our final reconstruction on the input image, final reconstruction, shading, normal, diffuse and specular.}
  \label{fig:catMakeup1}
\end{figure*}

\begin{figure*}
\hspace{35pt}Input\hspace{35pt} Overlay\hspace{65pt} Final\hspace{35pt} Shading\hspace{15pt} Normal\hspace{15pt} Diffuse\hspace{15pt} Specular \\
\centering
\includegraphics[width=0.9\linewidth]{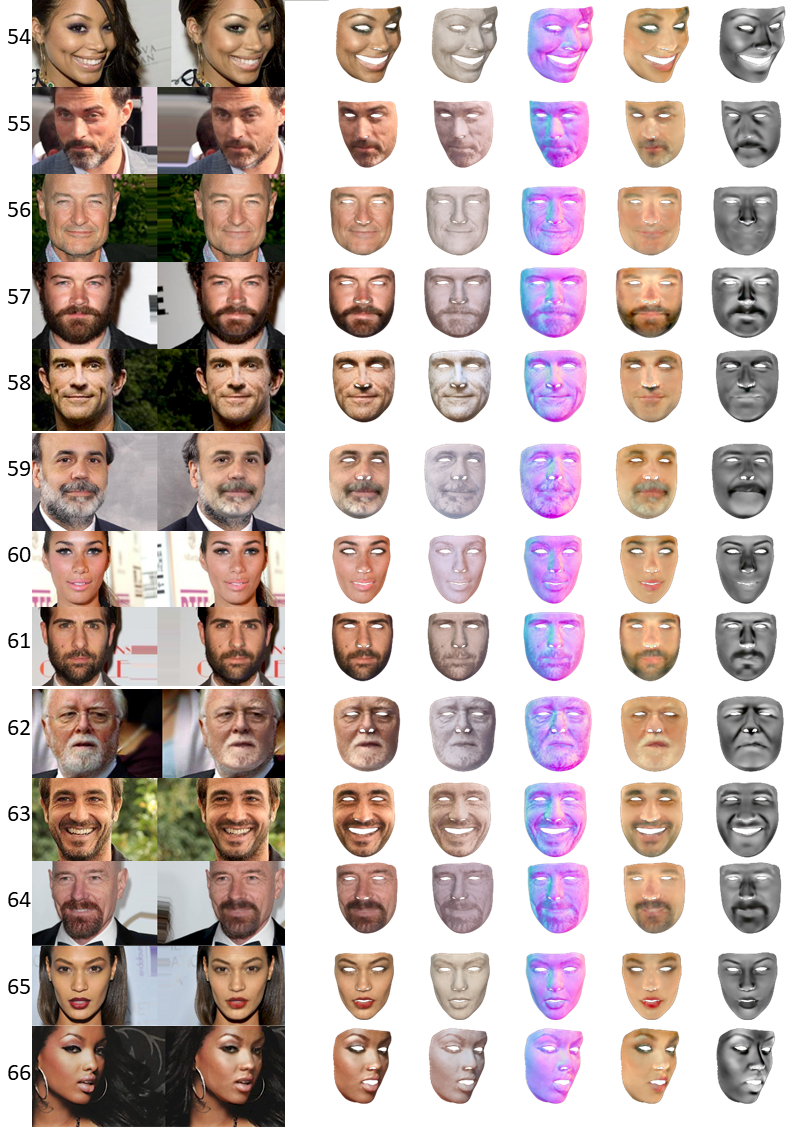}
  \caption{Results for subjects with make-up, beards and face props. From left to right: Input image, overlay of our final reconstruction on the input image, final reconstruction, shading, normal, diffuse and specular.}
  \label{fig:catMakeup2}
\end{figure*}

\begin{figure*}
\hspace{35pt}Input\hspace{35pt} Overlay\hspace{65pt} Final\hspace{35pt} Shading\hspace{15pt} Normal\hspace{15pt} Diffuse\hspace{15pt} Specular \\
\centering
\includegraphics[width=0.89\linewidth]{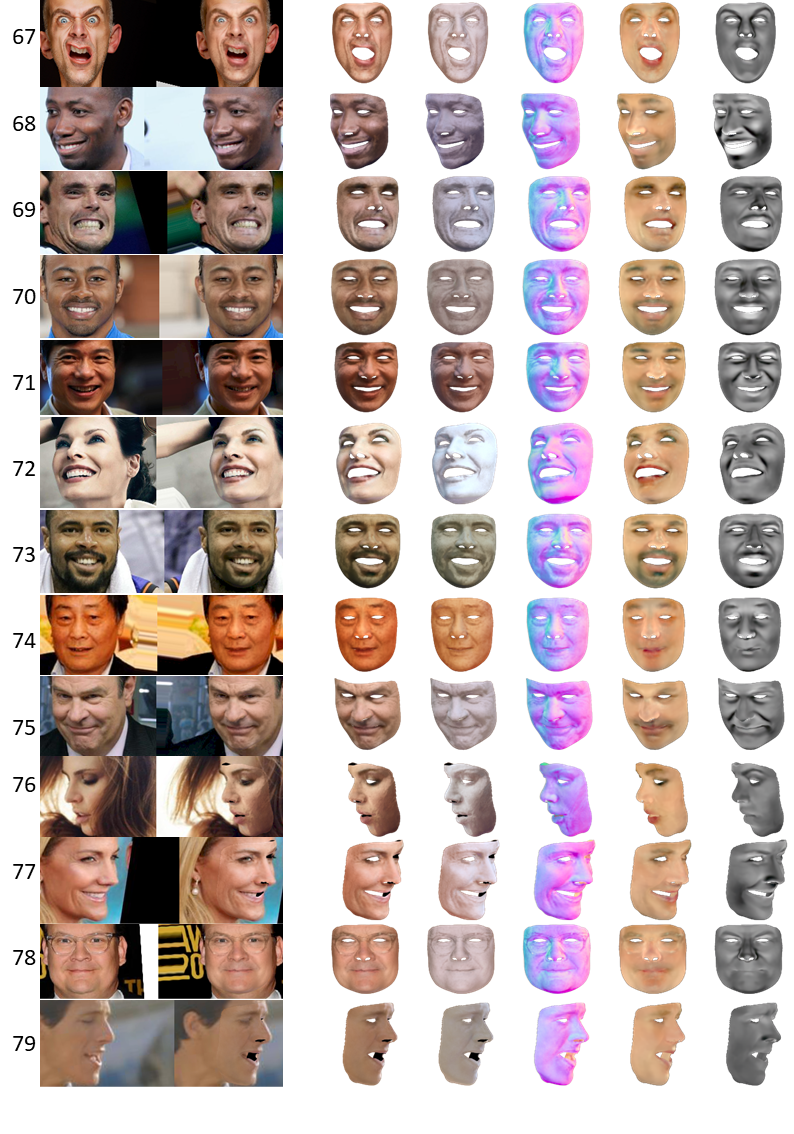}
  \caption{Results for subjects with different ethnicity, skin color, difficult expression and challenging head pose. From left to right: Input image, overlay of our final reconstruction on the input image, final reconstruction, shading, normal, diffuse and specular.}
  \label{fig:catDiv1}
\end{figure*}

\begin{figure*}
\hspace{35pt}Input\hspace{35pt} Overlay\hspace{65pt} Final\hspace{35pt} Shading\hspace{15pt} Normal\hspace{15pt} Diffuse\hspace{15pt} Specular \\
\centering
\includegraphics[width=0.9\linewidth]{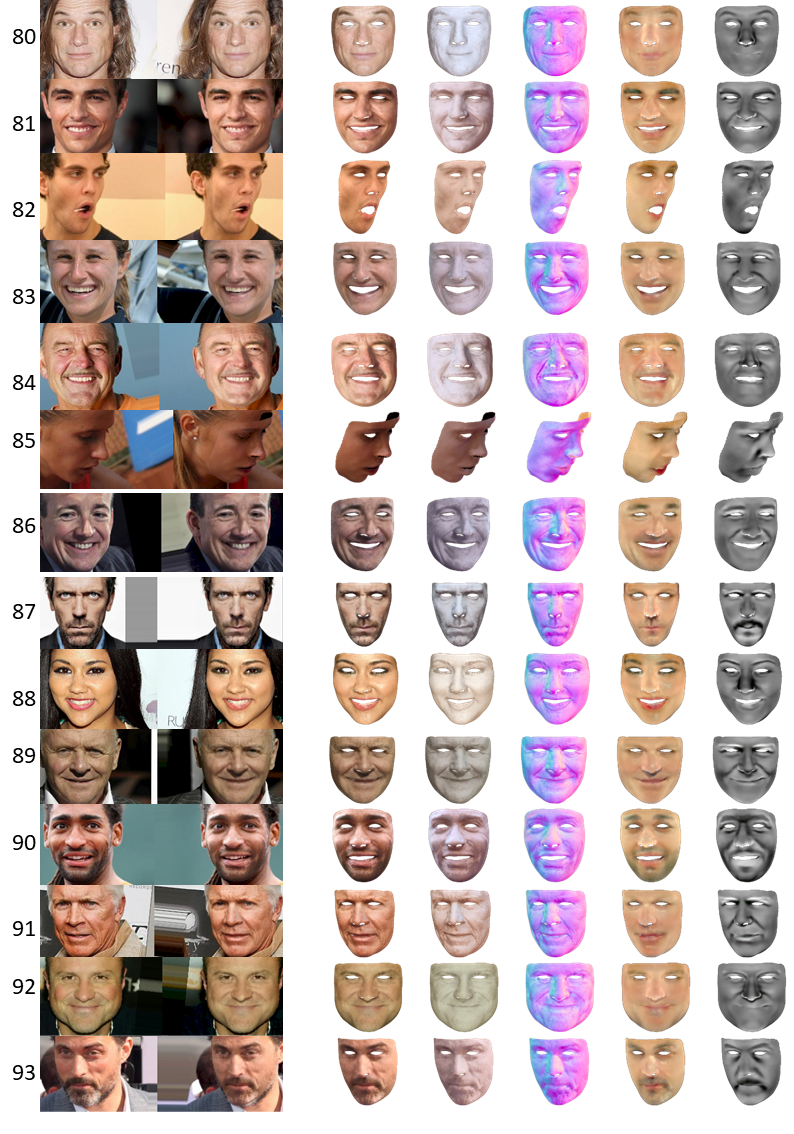}
  \caption{Results for subjects with different ethnicity, skin color, difficult expression and challenging head pose. From left to right: Input image, overlay of our final reconstruction on the input image, final reconstruction, shading, normal, diffuse and specular.}
  \label{fig:catDiv2}
\end{figure*}

\begin{figure*}
\hspace{35pt}Input\hspace{35pt} Overlay\hspace{65pt} Final\hspace{35pt} Shading\hspace{15pt} Normal\hspace{15pt} Diffuse\hspace{15pt} Specular \\
\centering
\includegraphics[width=0.89\linewidth]{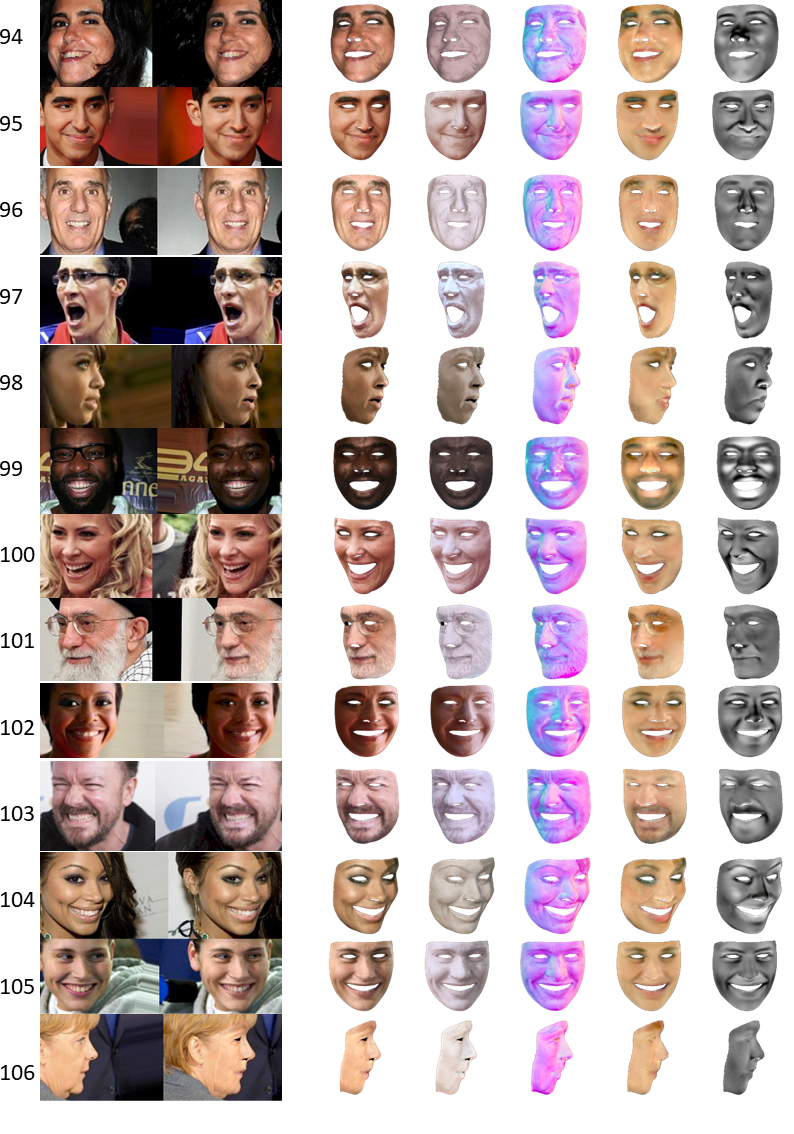}
  \caption{Results for subjects with different ethnicity, skin color, difficult expression and challenging head pose. From left to right: Input image, overlay of our final reconstruction on the input image, final reconstruction, shading, normal, diffuse and specular.}
  \label{fig:catDiv3}
\end{figure*}

\begin{figure*}
\hspace{35pt}Input\hspace{35pt} Overlay\hspace{65pt} Final\hspace{35pt} Shading\hspace{15pt} Normal\hspace{15pt} Diffuse\hspace{15pt} Specular \\
\centering
\includegraphics[width=0.89\linewidth]{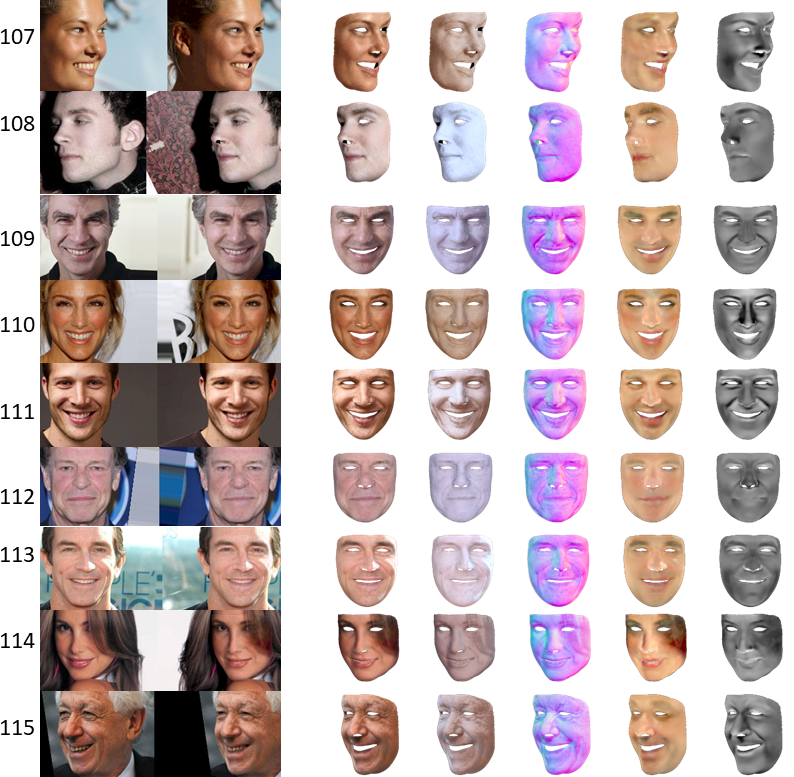}
  \caption{Results for subjects with different ethnicity, skin color, difficult expression and challenging head pose. From left to right: Input image, overlay of our final reconstruction on the input image, final reconstruction, shading, normal, diffuse and specular.}
  \label{fig:catDiv4}
\end{figure*}

\begin{itemize}
    \item Figure~\ref{fig:catOld1} and Figure~\ref{fig:catOld2} show reconstruction for subjects with challenging details on the face. 
    \item Figure~\ref{fig:catLight} show reconstruction for subjects under challenging lighting conditions. 
    \item Figure~\ref{fig:catMakeup1} and Figure~\ref{fig:catMakeup2} show reconstruction for subjects with make-up, beards and face props.
    \item Figures~\ref{fig:catDiv1}, \ref{fig:catDiv2}, \ref{fig:catDiv3} and  \ref{fig:catDiv4}  show reconstruction for various subjects with different ethnicity, skin color, difficult expression, challenging head pose. 
\end{itemize}

\begin{figure*}
\centering
\includegraphics[width=0.89\linewidth]{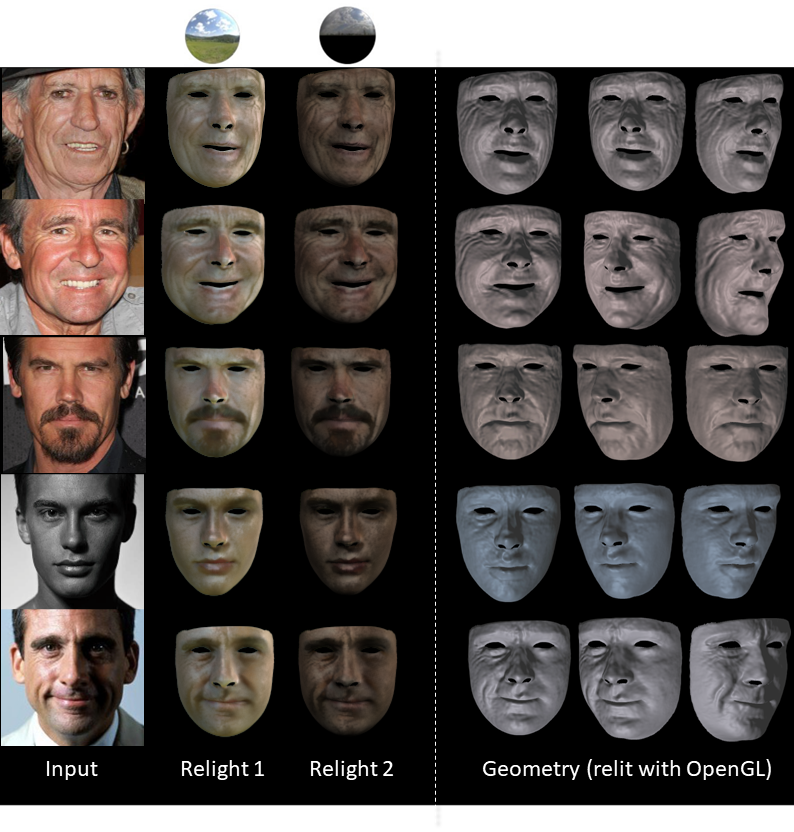}
  \caption{Our robust face attributes estimation gives explicit control over these attributes and allow for relighting even for subjects under challenging lighting conditions (last two subjects). The last three columns show the estimated geometry by our method for each subject rendered with OpenGL under different viewing angles.}
  \label{fig:relighting}
\end{figure*}

For all these subjects, our method generalizes very well and achieves appealing 3D reconstruction with high fidelity compared to the input image (first column vs  second column). It also successfully estimates meaningful  face attributes with faithful separation between reflectance and detailed geometry. Even under challenging lighting conditions, our method  estimates plausible face reconstruction and produces shadow-free maps. In the supp. video, we show animated reconstruction with relighting and also reconstruction from video sequence. 
\\
Figure~\ref{fig:relighting} shows relighting of different subjects under novel lighting conditions (second and third column). Because our method can successfully estimates shadow-free face attributes, and faithfully separates reflectance from detailed geometry, all this allow our method to perform relighting even for subjects under challenging light (last two subjects in Figure \ref{fig:relighting}). The last three columns in Figure~\ref{fig:relighting} show the estimated detailed geometry by our method rendered with OpenGL under different viewing angles.

\begin{figure*}
\centering
\includegraphics[width=0.89\linewidth]{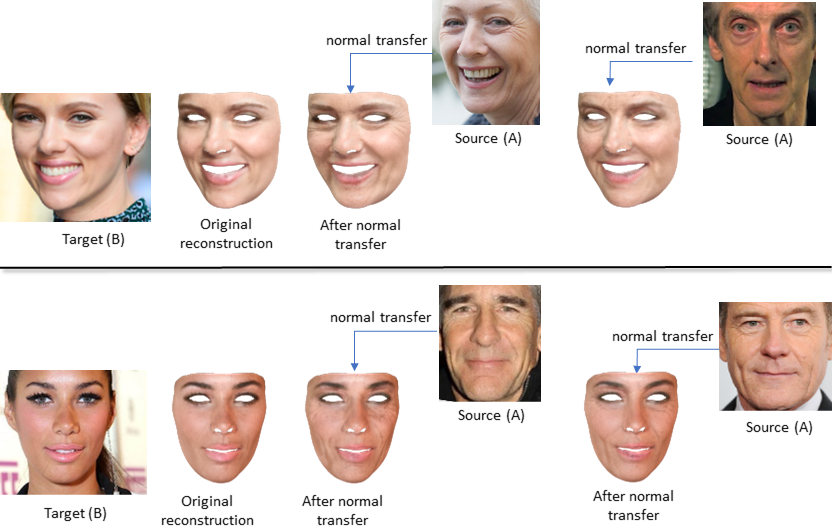}
  \caption{Face attributes edition: Our robust face attributes estimation allow for practical applications such as normal transfer from source (A) to target (B) which leads to aging/de-aging (here we show aging). Please note the wrinkles/folds that appears on the aged face.}
  \label{fig:app}
\end{figure*}

Another practical applications for our method are face attribute(s) editing,  attribute(s) transfer, aging and de-aging... We show in Figure \ref{fig:app} an 'Aging' application that consists simply on transferring the estimated normal map from source (A) to target (B). 

\end{document}